\documentclass[journal]{IEEEtran}
\usepackage{times}
\usepackage{latexsym}
\usepackage{url}
\usepackage{algorithm}
\usepackage{algcompatible}
\usepackage[noend]{algpseudocode}
\usepackage{amssymb}
\usepackage{multirow}
\usepackage{verbatim}
\usepackage{scrextend}
\usepackage{float}
\usepackage{subfig}
\usepackage{array}
\usepackage{mathtools}
\newcommand*\rot{\rotatebox{90}}
\usepackage{tablefootnote}
\newcommand{\slfrac}[2]{\left.#1\middle/#2\right.}

\newcolumntype{C}[1]{>{\centering\let\newline\\\arraybackslash\hspace{0pt}}m{#1}}
\usepackage{tikz}
\usetikzlibrary{shapes,arrows}
\usepackage{amsmath,bm,times}
\usepackage{verbatim}
\usepackage{subfig}
\let\labelindent\relaxs
\usepackage{enumitem}

\DeclarePairedDelimiter{\ceil}{\lceil}{\rceil}
\usepackage{float}
\usetikzlibrary{positioning}
\usepackage{tkz-graph}

\usetikzlibrary{decorations.pathreplacing,intersections}
\usepackage{wrapfig}
\usepackage{pbox}
\usepackage{colortbl}
\usepackage{xcolor,soul,framed}
\colorlet{shadecolor}{yellow}
\usepackage{eqparbox}
\usepackage{url}
\usepackage[square,numbers]{natbib}
\usepackage{wrapfig}
\usepackage{pifont}
\usepackage{amssymb}

\newcommand{\name}{\texttt{MSH-COMICS}}%
\newcommand{\datasetname}{\texttt{MaSaC}}%

\newcommand{\xmark}{\ding{55}}%

\begin{document}

\title{Multi-modal Sarcasm Detection and Humor Classification in Code-mixed Conversations}

\author{Manjot Bedi$^*$, Shivani Kumar$^*$, Md Shad Akhtar, and Tanmoy Chakraborty\\ 
  \textit{Dept. of CSE, IIIT-Delhi, India} \\
  {\tt \{manjotb, shivaniku, shad.akhtar, tanmoy\}@iiitd.ac.in} \\
  \thanks{$^*$Equal contribution.}
} 

\markboth{IEEE TRANSACTIONS ON AFFECTIVE COMPUTING. 
}{ \MakeLowercase{\textit{et al.}}:}

% ====================================================================
\maketitle

\begin{abstract}
Sarcasm detection and humor classification are inherently subtle problems, primarily due to their dependence on the contextual and non-verbal information. Furthermore, existing studies in these two topics are usually constrained in non-English languages such as Hindi, due to the unavailability of qualitative annotated datasets. In this work, we make two major contributions considering the above limitations: (1) we develop a Hindi-English code-mixed dataset, \datasetname\footnote{\datasetname\ can be vaguely pronounced as \textit{Mazaak} (Joke) in Hindi.}, for the multi-modal sarcasm detection and humor classification in conversational dialog, which to our knowledge is the first dataset of its kind; (2) we propose \name\footnote{\name\ is short for \textbf{M}ulti-modal \textbf{S}arcasm Detection and \textbf{H}umor Classification in \textbf{CO}de-\textbf{MI}xed \textbf{C}onversation\textbf{S}.}, a novel attention-rich neural architecture for the utterance classification. We learn efficient utterance representation utilizing a hierarchical attention mechanism that attends to a small portion of the input sentence at a time. Further, we incorporate dialog-level contextual attention mechanism to leverage the dialog history for the multi-modal classification. We perform extensive experiments for both the tasks by varying multi-modal inputs and various submodules of \name. We also conduct comparative analysis against existing approaches. We observe that \name\ attains superior performance over the existing models by $>$1 F1-score point for the sarcasm detection and 10 F1-score points in humor classification. We diagnose our model and perform thorough analysis of the results to understand the superiority and pitfalls. 
\end{abstract}


\IEEEpeerreviewmaketitle

\section{Introduction}\label{sec:intro}
Unlike traditional sentiment \cite{Turney:2002,Panget.al2005,taffc:2016:sentiment,senti:lstm:cambria:2018} or emotion \cite{ekman1999,taffc:2019:multitask:acoustic:emotion,akhtar:all:in:one:affective,akhtar:cim:emotion:2020,kumar2021discovering} classification, 
sarcasm or humor detection in a standalone textual input (e.g., a tweet or a news headline) is a non-trivial task due to its below-the-surface semantics. Most of the time, the surface-level words carry sufficient cues in the text to detect the expressed sentiment or emotion. However, sarcastic or humorous inputs do not offer such simplistic information for classification. Instead, the expressed semantic information  in sarcastic or humorous inputs often have dependency on the context of the text, and it is important to leverage the contextual information for the identification task. Moreover, in many cases, the presence of multi-modal signals, such as visual expression, speech pattern, etc., provide auxiliary but crucial cues for sarcasm or humor detection. At times, they are the only cues that support a sarcastic/humorous expression. For example, it is extremely difficult (or nearly impossible) to detect sarcasm in the text `\textit{Thanks for inviting me!}' without any context or other information. However, the same is less challenging if multi-modal signals accompany the text (e.g., the disgusting facial expression, gaze movement, or intensity/pitch of the voice while uttering the text) or the context (e.g., the text was preceded by a dispute/argument/insult).

A conversational dialog records the exchange of utterances among two or more speakers in a time series fashion. Thus, it offers an excellent opportunity to study the sarcasm or humor in a context. A few previous attempts \cite{cai-etal-2019-multi,castro-etal-2019-towards:mustard,suyash:multimodal:sarcasm:ijcnn:2020} on sarcasm classification involved multi-modal information in a conversation to leverage the context and extract the incongruity between the surface and expressed semantics. Similarly, many studies \cite{hasan-etal-2019-ur, humor:text:audion:lrec:2016} employed images and visual frames along with the text to detect humor. Surveys on multi-modal analysis \cite{multimodal1,ghosal-EtAl:2018:EMNLP,zadeh2018acl,akhtar:naacl:2019:multitask:multimodal,akhtar:multimodal:tkdd:2020,taffc:2019:audio:visual:emotion} reveal two prime objectives while handling multi-modal contents: (a) to leverage the distinct and diverse information offered by each modality, and (b) to reduce the effect of noise among the multi-modal information sources. 
% Code-mix example
\begin{table}[t]
    \centering
    \begin{tabular}{l|p{23em}}
        \multicolumn{2}{c}{\bf Code-mixed} \\ \hline
        Original: & \textit{Sachin \textcolor{red}{ne} 21 \textcolor{blue}{years} \textcolor{red}{pehle apna} \textcolor{blue}{debut match} \textcolor{red}{khela tha}.} \\
        Translation: & Sachin played his debut match 21 years ago. \\ \hline
       \multicolumn{2}{c}{} \\
    
       \multicolumn{2}{c}{\bf Code-switched} \\ \hline
        Original: & \textit{\textcolor{red}{Agle hafte meri garmi ki chuttiyan shuru hone wali hain}. \textcolor{blue}{I am planning to go to Europe during my vacation.}} \\
        Translation: & My summer vacation is starting next week. I am planning to go to Europe during my vacation. \\ \hline
    \end{tabular}
    \caption{Examples for the code-mixed and code-switched inputs. Blue-colored text represents English words and red-colored text signifies Hindi words.}
    \label{tab:code-mix:code-switch:exm}
    \vspace{-3mm}
\end{table}
Usually, the solution to a natural language processing task handles only a single language. However, with the globalization of languages, many applications demand for solutions that can handle more than one language at a time. Thus, a new frontier of multi-lingual processing has emerged. India is a multi-lingual country, and a vast population are comfortable with more than one language. Their comfort is apparent in the regular usage of words from multiple languages to form a single sentence in both writing and speaking. For example, the text `\textit{Sachin ne 21 years pehle apna debut match khela tha.}' (`{\em Sachin played his debut match twenty one years ago.}') has three English words (i.e., `years', `debut', and `match') and one named-entity (i.e., `Sachin'), while the rest of the words are part of romanized Hindi language. Similarly, it is common to switch languages for the consecutive sentences as well. For example, two sentences in Table \ref{tab:code-mix:code-switch:exm} are in two different languages -- not only their words are in different languages, but also they follow language-specific syntactic structure. These two variants are usually termed as the {\em code-mixed} and {\em code-switched} inputs, respectively.   

Though the code-mixed and code-switched inputs are natural in a multi-lingual culture, they offer a significant challenge in the automatic processing of such text. The foremost task in handling code-mixed input is the language identification of each word. Dictionary-based lookup is a trivial solution to identify language-specific words; however, the complexity escalates when a token (in transliterated form) is a valid word in more than one language. For example, the word `\textit{main}' has the meaning `important' in English, while it also means `\textit{I}' in Hindi. Once the language is identified for each word, literature suggests language-specific processing for the downstream tasks in a trivial setup. Recently, the quest of handling multi-lingual inputs in a deep neural network architecture has paved the way for the development of more sophisticated multi-lingual/cross-lingual word representation techniques \cite{conneau2017word,lample2017unsupervised}.

Most of the existing datasets for the multi-modal sarcasm and humor detection involve only monolingual data (primarily English). To explore the challenges of code-mixed scenarios, in this paper, we introduce \datasetname, a new multi-modal contextual sarcasm and humor classification dataset in English-Hindi code-mix environment. \datasetname\ comprises $\sim$1,200 multi-party dialogs extracted from a popular Indian television show `\textit{Sarabhai vs. Sarabhai}'\footnote{\url{https://www.imdb.com/title/tt1518542/}}. It contains $\sim$15,000 utterance exchanges (primarily in Hindi) among the speakers. We manually analyze all the utterances and mark the presence/absence of sarcasm and humor for each of them (c.f. Section \ref{sec:dataset} for detailed description).

To evaluate \datasetname\ dataset, we propose \name, a multi-modal hierarchical attention framework for the utterance classification in conversational dialogs. At first, we encode the textual utterance representation using a hierarchy of localized attention over the tokens in a sentence. In the next step, we learn the modality-specific dialog sequence using LSTM \cite{lstm1997} layers. Further, to leverage the contextual information, we employ three attention mechanisms that learn the importance of preceding utterances with respect to each of the textual, acoustic, and textual+acoustic modalities. Since one of the prime concerns in multi-modal analysis is to counter the presence of noise among modalities, we employ a simple gating mechanism that aims to filter the noise in accordance with the interactions among the modalities. Finally, we utilize the filtered representations for the sarcasm and humor classification.

Experimental results suggest significant performance for both the sarcasm and humor classification tasks. We also evaluate \datasetname\ on the existing multi-modal contextual sentence classification systems. The comparative study reveals that \name\ yields superior performance compared to the baselines for both the tasks.     

The contributions of the current work are as follows:
\begin{itemize}[leftmargin=*]
    \item We develop \datasetname, a qualitative multi-modal dataset for the sarcasm detection and humor classification. 
    \item We propose a novel architecture for the multi-modal contextual sentence classification.
    \item We provide strong baselines for the two tasks on the proposed dataset.
    \item We report detailed analysis of the experimental results and the reported errors. 
    \item Through our developed \datasetname\ dataset, we offer an opportunity to the community to carry forward the research on the code-mixed environment in Indian context.\footnote{\url{https://github.com/LCS2-IIITD/MSH-COMICS.git}}  
\end{itemize}

The rest of the paper is organized as follows: We formulate the problem in Section \ref{sec:problem}, while Section \ref{sec:lit} reports relevant related works. In Section \ref{sec:method}, we describe our proposed model. We elaborate on the dataset development in Section \ref{sec:dataset}. Section \ref{sec:experiment} presents our experimental results and necessary analysis. Finally, we conclude the paper  in Section \ref{sec:con}.

\section{Problem Definition}
\label{sec:problem}
Sarcasm is defined as an expression meant to criticize, taunt, or hurt someone's feeling in a sober and explicitly non-disrespectful manner. On the other hand, humorous statements aim to incite amusing or comic feelings with the intention to make their audience laugh.
A light-hearted sarcastic statement which does not offend the target can be interpreted as humorous. However, it is important to note that all sarcastic statements may not be amusing, whereas a humorous expression is always intended to amuse the listeners.

\begin{figure}
    \centering
    \includegraphics[width=0.7\columnwidth]{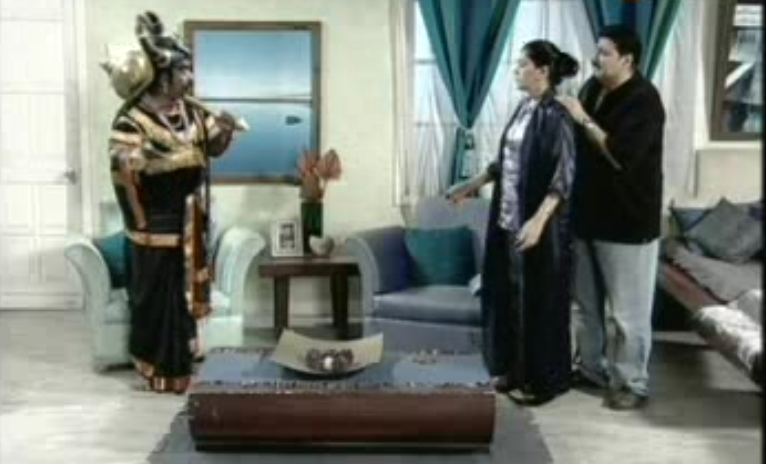}
    \resizebox{\columnwidth}{!}
    {%
        \begin{tabular}{ll}
            \textit{Kyun? Mene to apna bhensa \textcolor{blue}{building} ke bahar park kiya hai...} & [Humour] \\
            (Why? I have parked my bull outside the building...) &  \\
        \end{tabular}
    }
    \caption{An example frame highlighting the irrelevant visual content considering the humour (or sarcasm) prediction, and the model may defile the learning process by attending to irrelevant contents (or noise).}
    \label{fig:frames}
\end{figure}

In the current work, our objective is to identify all the instances of sarcastic or humorous utterances in a multi-speaker conversational dialog. 
Given a sequence of utterance $U = (u_1, u_2, ..., u_n)$ in a dialog video, we wish to classify each utterance into -- (i)  \textit{sarcastic} or \textit{non-sarcastic}, and (ii)  \textit{humorous} or \textit{non-humorous}. Each utterance $u_i$ has multiple representations corresponding to the available modalities, i.e., visual frames of the utterance $u_i^{V}$, acoustic signals of the utterance $u_i^{A}$, and the utterance transcripts $u_i^{T}$. In our study, we do not account for the visual frames while learning the model. A valid explanation for leaving out the visual modality is due to the presence of multiple actors in a frame, and most of them do not offer any constructive assistance to the model. We argue that the inclusion of the visual frames in the model would defile the learning process by attending to irrelevant content (or noise).To support our claim, in Figure \ref{fig:frames}, we show one of many such scenarios.
Therefore, we employ only the textual and acoustic features in our model.

\section{Related Work}
\label{sec:lit}
In this section, we present a survey of the literature on the sarcasm detection and humor classification focusing on the following three dimensions -- context, multi-modality, and Indian languages.

\textbf{Sarcasm Detection:} Sarcasm detection is an interesting as well as a challenging task. It has gained significant attention in the last few years \cite{kreuz-caucci:2007:sarcasm:lexical,davidov:sarcasm:semisupervised:2010,tsur:sarcasm:2010,joshi:sarcsam:incongruity:2015,sarcasm:twitter:pattern:ieee:access:2016,joshi:sarcasm:survey:2017}. Earlier work on sarcasm detection involves investigation on the lexical aspects of the text expressing sarcasm \cite{kreuz-caucci:2007:sarcasm:lexical}. More specifically, the authors studied the influence of adjectives, adverbs, interjections, and punctuation marks in sarcasm detection, and showed that their presence have positive correlation (though small) with the sarcastic text. Tsur et al. \cite{tsur:sarcasm:2010} proposed a semi-supervised approach for sarcasm discovery in Amazon product reviews. The authors employed punctuation and pattern-based features to classify the unseen samples using a kNN classifier. A similar study on tweet was proposed in \cite{davidov:sarcasm:semisupervised:2010}.     
Other works claimed the presence of sentiment shift or the contextual incongruity to be an important factor in accurate sarcasm prediction \cite{joshi:sarcsam:incongruity:2015}.       
Son et al. \cite{sarcasm:soft:attention:ieee:access:2019} proposed a hybrid Bi-LSTM and CNN based neural architecture for the sarcasm detection. 

Most of the above studies involve sarcasm discovery in the standalone input - which are reasonably adequate for the sentence with explicit sarcastic clues. However, for the implicit case, more often than not, the context in which the sarcastic statement was uttered is of utmost importance \cite{joshi:sarcasm:history:2015,ghosh-veale-2017-magnets,hazarika-etal-2018-cascade}. Joshi et al. \cite{joshi:sarcasm:history:2015} exploited the historical tweets of a user to predict sarcasm in his/her tweet. They investigated the sentiment incongruity in the current and historical tweets, and proposed it to be a strong clue in the sarcasm detection. In another work, Ghosh et al. \cite{ghosh-etal-2017-role} employed an attention-based recurrent model to identify sarcasm in the presence of a context. The authors trained two separate LSTMs-with-attention for the two inputs (i.e., sentence and context), and subsequently, combined their hidden representations during the prediction. The availability of context was also leveraged by \cite{ghosh-veale-2017-magnets}. The authors learned a CNN-BiLSTM based hybrid model to exploit the contextual clues for sarcasm detection. Additionally, they investigated the psychological dimensions of the user in sarcasm discovery using $11$ emotional states (e.g., \textit{upbeat}, \textit{worried}, \textit{angry}, \textit{depressed}, etc.).           

Although a significant number of studies on sarcasm detection have been conducted in English, only a handful attempts have been made in Hindi or other Indian languages \cite{DBLP:journals/corr/abs-1805-11869,sarcasm:hindi:2017}. One of the prime reasons for limited works is the absence of sufficient dataset on these languages. Bharti et al. \cite{sarcasm:hindi:2017} developed a sarcasm dataset of 2,000 Hindi tweets. For the baseline evaluation, they employed a rule-based approach that classifies a tweet as sarcastic if it contains more positive words than the negative words, and vice-versa. In another work, Swami et al. \cite{DBLP:journals/corr/abs-1805-11869} collected and annotated more than 5,000 Hindi-English code-mixed tweets. They extracted n-gram and various Twitter-specific features to learn SVM and Random Forest classifiers. Though the dataset proposed by Swami et al. \cite{DBLP:journals/corr/abs-1805-11869} and \datasetname\ involve Hindi-English code-mixed inputs, they differ on the contextual dimensions, i.e., the instances in their dataset are standalone and do not have any context associated with them, whereas, the sarcastic instances in \datasetname\ are a part of the conversational dialog. Moreover, \datasetname\ also includes multi-modal information for each dialog.          

Recently, the focus on sarcasm detection has shifted from the text-based uni-modal analysis to the multi-modal analysis \cite{cai-etal-2019-multi,castro-etal-2019-towards:mustard}. Cai et al. \cite{cai-etal-2019-multi} proposed a hierarchical fusion model to identify the presence of sarcasm in an image in the pretext of its caption. The authors exploited the incongruity in the semantics of the two modalities as the signals of sarcasm. Another application of the multi-modal sarcasm detection is in the conversational dialog system. During the conversation, it is crucial for a dialog agent to be aware of the sarcastic utterances and respond accordingly. Castro et al. \cite{castro-etal-2019-towards:mustard} developed a multi-speaker conversational dataset for the sarcasm detection. For each sarcastic utterance in the dialog, the authors identified a few previous utterances as the context for sarcasm. The dataset developed in the current work is on the similar line except two major differences: (a) \datasetname\ contains Hindi-English code-mixed utterances, which is the first dataset of its kind; and (b) instead of defining the explicit context, we let the model learn the appropriate context during training.

\begin{figure*}[t]
    \centering
    \includegraphics[width=0.7\textwidth]{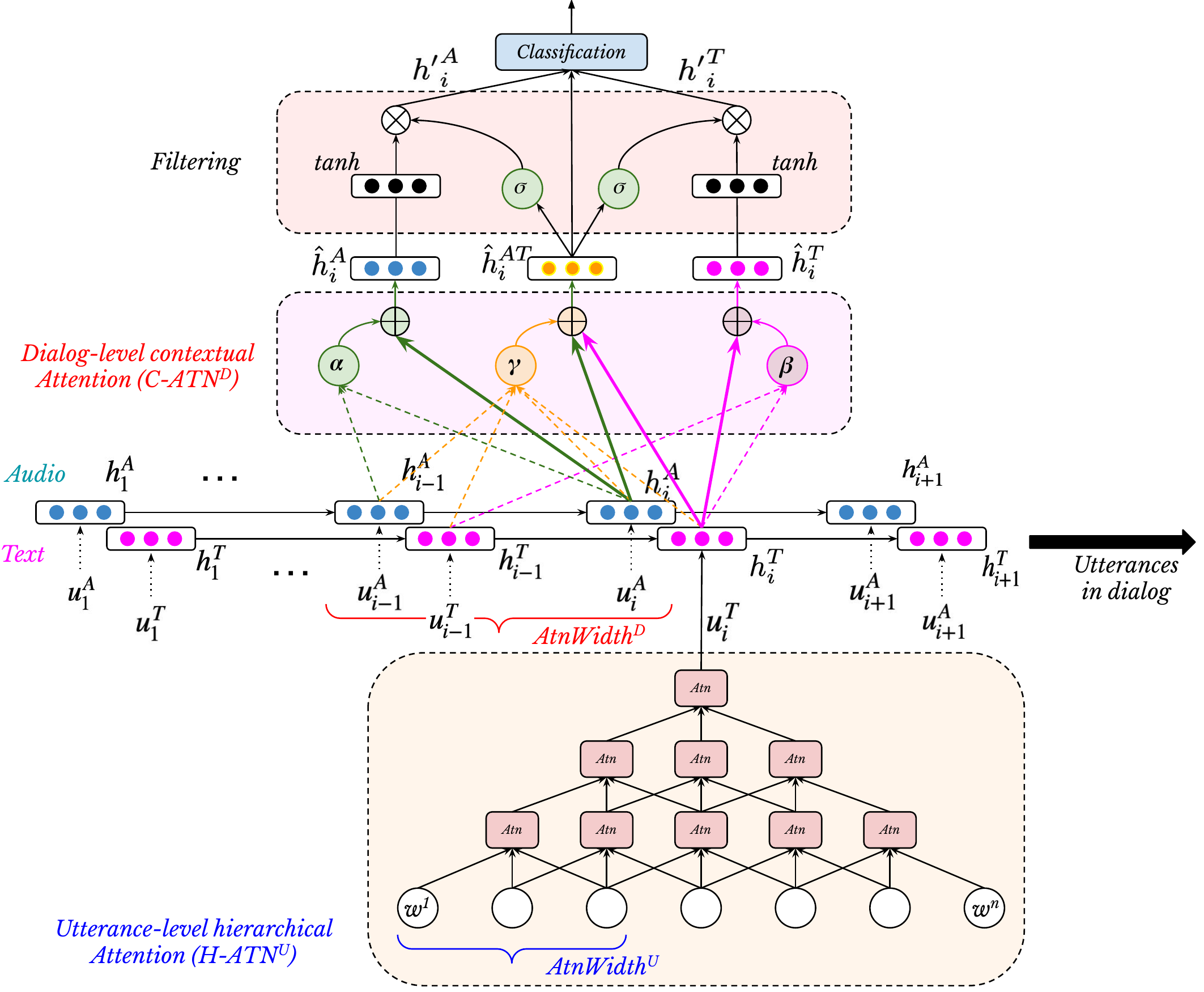}
    \caption{System architecture of \name. Each instance is a sequence of utterances in a conversational dialog, and the classification is performed for each utterance. \textit{H-ATN}$^U$ computes efficient textual representation for each utterance in the dialog. {\color{black} \textit{C-ATN}$^D$} learns attention weights of the contextual (preceding) utterances considering the acoustic ($\alpha$), textual ($\beta$), and cross-modal ($\gamma$) patterns. Filtering mechanism reduces the effect of noise in the learned representation of individual modality.}
    \label{fig:arch}
    \vspace{-5mm}
\end{figure*}

\textbf{Humor Detection:} Like sarcasm detection, computational humor analysis is a fascinating but subtle task in the domain of natural language processing. Recent literature suggests that contextual information plays an important role in computational humor detection \cite{humor:text:audion:lrec:2016,hasan-etal-2019-ur}. However, due to the complexity in processing the contextual information, many of the earlier studies aim to identify humorous contents in standalone text without consulting the context \cite{mihalcea-strapparava-2005-making, semeval:2017:humor:potash-etal, Yang2019, annamoradnejad2021colbert}. Their prime inputs are \textit{one-liners} or \textit{punchlines} - which usually have rich comic or rhetoric content to attract someone's attention. Though the strategy of detecting humor in standalone texts seem appealing, often the absence of context makes it extremely difficult (even for humans) to interpret the humorous content. Moreover, the textual form of the humorous contents are complemented with other crucial non-verbal signals such as animated voice, impersonation, funny facial expression, etc. This difference in acoustic features between humorous and non-humorous utterances is validated by Amruta et al. \cite{Humor:Prosody} Many researchers have exploited these meta-data for humor classification \cite{hasan-etal-2019-ur, humor:text:audion:lrec:2016, bertero2016predicting}. Hasan et al. \cite{hasan-etal-2019-ur} extended humor classification in punchlines by considering both the contextual and multi-modal information. The authors utilized Transformer's \cite{vaswani:transformer:attention:all:you:need:2017} encoder architecture to model the contextual information in addition to the memory fusion network \cite{mfn:zadeh:aaa1:2017} for combining the multi-modal signals. Bertero and Fung \cite{humor:text:audion:lrec:2016} relied on the text and acoustic features for contextual humor classification. Dario et al. \cite{bertero2016predicting} treated the humor classification task as sequence labelling and employed conditional random field to get the output. In the context of Indian languages, the study on humor classification, like any other NLP task, is limited. To the best of our knowledge, Khandelwal et al. \cite{khandelwal-etal-2018-humor} is one of the first studies that involve humor classification in Hindi-English code-mixed language. They developed a dataset of $\sim$3,500 tweets with almost equal number of humorous and non-humorous tweets. The authors bench-marked the dataset on SVM classifier using bag-of-word features. Sane et al. \cite{sane-etal-2019-deep} improved the state-of-the-art on the same dataset using neural models. In comparison with \datasetname, the dataset of Kandelwal et al. \cite{khandelwal-etal-2018-humor} lacks both the contextual as well as multi-modal information. Furthermore, \datasetname\ has significantly more number of instances, and annotations for two tasks, i.e., sarcasm and humor detection.          

\section{Methodology}
\label{sec:method}
In this section, we describe \name, our proposed system for sarcasm and humor classification. Figure \ref{fig:arch} presents a high-level architectural overview of \name. It takes a sequence of utterances (a dialog) as input and produces a corresponding label for each utterance.

To learn the context of the dialog, we employ two LSTMs on top of textual ($u_i^T$) and acoustic ($u_i^A$) representations of dimensions $d^T$ and $d^A$, respectively. 
\begin{eqnarray}
h_i^A = LSTM^A(u_i^A, h_{i-1}^A) \\
h_i^T = LSTM^T(u_i^T, h_{i-1}^T)
\end{eqnarray}
where $h_i^A \in \mathbb{R}^{d^A}$ and $h_i^T \in \mathbb{R}^{d^T}$ are the learned hidden representations for acoustic and textual modalities, respectively. The textual representation ($u_i^T$) is computed through an application of the utterance-level hierarchical attention module (discussed later), whereas, the acoustic representation ($u_i^A$) is obtained through an audio processing tool, Librosa \cite{brian_mcfee-proc-scipy-2015} (c.f. Section \ref{sec:dataset:feature:ex}). \\

\noindent \textbf{Dialog-level contextual attention (\textit{C-ATN}$^D$):} Subsequently, we employ three separate attention modules that compute attention weights (i.e., $\alpha, \beta,$ and $\gamma$) of the contextual (preceding) utterances considering the acoustic pattern, textual pattern, and cross-modality pattern. 
\begin{eqnarray}
\alpha_i & = & \frac{\exp(h_i^A)}{\sum_{j=1}^{i} \exp(h_j^A)} \\
\beta_i & = & \frac{\exp(h_i^T)}{\sum_{j=1}^{i} \exp(h_j^T)} \\
\gamma_i & = & \frac{\exp(h_i)}{\sum_{j=1, X \in (A, T)}^{i} \exp(h_j^X)} \end{eqnarray}

These attention weights signify the importance of contextual utterances $u_1,..., u_{i-1}$ for the classification of utterance $u_i$. Therefore, we compute the mean of the contextual attended vectors for each hidden representation $h_i$. Further, we utilize the residual skip connection \cite{he2016deep} to form the final attended representations $\hat{h}_i^A$, $\hat{h}_i^T$, and $\hat{h}_i^AT$ corresponding to the acoustic, textual, and cross-modal attention modules, respectively as follows: 
\begin{eqnarray}
\hat{h}_i^A & = & \sum_k^i \alpha_k h_k^A / i \oplus h_i^A \\
\hat{h}_i^T & = & \sum_k^i \beta_k h_k^T / i \oplus h_i^T \\
\hat{h}_i^{AT} & = & \sum_{k, X \in (A, T)}^i \gamma_k h_k^{X} / 2i \oplus h_i^A \oplus h_i^T
\end{eqnarray}
where $\oplus$ is the concatenation operator. Collectively, we term these three attention modules as the dialog-level contextual attention module \textit{C-ATN}$^D$, i.e., \textit{C-ATN}$^D = [\hat{h}_i^A, \hat{h}_i^T, \hat{h}_i^{AT}]$. In the subsequent steps, we consume these representations for the final classification.\\

\noindent \textbf{Filtering:} Prior to feeding these representations to the fully-connected layers, we incorporate a noise filtering mechanism \cite{arevalo:2017:gmu:gated} to enhance the representation for each modality. The intuition behind the filtering mechanism is to learn the interaction among the available modalities, which has not been incorporated in the model so far, and subsequently, filter the noise in correspondence with the other modalities. We argue that the filtering mechanism provides assistance to the model to pass only the relevant features such that the filtered representations of different modalities can complement each other in retaining the diverse and distinct features. For each modality, we implement filtering as follows:
\begin{eqnarray}
{h'}_i^A & = & tanh(\hat{h}_i^A) \cdot \sigma(\hat{h}_i^{AT}) \\
{h'}_i^T & = & tanh(\hat{h}_i^T) \cdot \sigma(\hat{h}_i^{AT})
\end{eqnarray}
where $\sigma(\cdot)$ refers to the sigmoid function and is learned during the training. Since $\sigma(\cdot)$ lies in the range [0, 1], it controls the amount of information that can pass through the filter, i.e., a value close to 0 signifies extremely irrelevant information and is blocked, whereas, for a value approaching 1, all the information can be forwarded to the upper layers. Finally, we take the filtered representations along with the cross-modal attended vector for the final classification.\\

\noindent \textbf{Hierarchical attention module \textit{H-ATN}$^U$:} One of the crucial aspects of a deep neural architecture for any natural language processing task is the efficient input representation. Literature suggests the availability of many techniques to obtain an efficient sentence vector from the word-level embeddings, e.g., mean of constituent word embeddings, the last time-step representation in a recurrent layer, etc. However, a significant challenge in such approaches is to reduce the effect of irrelevant words and to find relations among the far apart words in the sentence. In this paper, we propose a hierarchical attention module \textit{H-ATN}$^U$ to learn the significance of constituent words in the final sentence vector. We apply a series of localize attentions, each one attending to a small portion of the sentence. For example, \textit{AtnWidth}$^U$=3 signifies that each attention mechanism attends to a sequence of three words only, and a context vector is obtained by taking a mean of the attended vectors followed by a linear layer with ReLU activation. As a consequence, we obtain \textit{N - AtnWidth}$^U$ + 1 context vectors at the first hierarchical level \textit{l}=1, where \textit{N} is the number of words in a sentence. Similarly, we apply localized attentions at the second hierarchical level \textit{l}+1, i.e., on \textit{N - AtnWidth}$^U$ + 1 context vectors. Following this process, we compute localized attention for $\ceil*{\frac{N-1}{\text{\textit{AtnWidth}}^U - 1}}$ hierarchical levels, and at the final level, we obtain a single context vector representing the entire sentence. It is to be observed that, as we go higher in the hierarchy, \textit{H-ATN}$^U$ attends to a wider sequence of words, thus offers a mechanism to extract long-term relations.         
We formulate the utterance-level hierarchical attention mechanism in Algorithm \ref{alg:hierarchical:atn}.

% algorithm
\begin{algorithm}[t]
\caption{Utterance-level Hierarchical Attention (\textit{H-ATN}$^U$)}\label{alg:hierarchical:atn}
\small
\begin{algorithmic}
\Procedure{\textit{H-ATN}$^U$}{$[w_1,., w_N] = W, X$ = \textit{AtnWidth}$^U$}
    \For{$k \in 1,..., N$}
        \State $CV_{(0,k)} = ReLU(w_{l})$
    \EndFor
    \State $M = ceiling(\slfrac{(N-1)}{(X - 1)})$ 
    
    \For{$l \in 1,..., M$}
        \State $Q = N - (l * X) + l$ 
        \For{$k \in 1,...,Q$}         
            \State $\zeta_{l,k} = Attention (CV_{(l-1, k)}, ..., CV_{(l-1, k + X - 1)})$
            \State $\phi_{l,k} = \slfrac{(\sum_{i}^{X} \zeta_{l,k} \cdot CV_{(l-1, k+i-1)})} {X}$
            \State $CV_{(l,k)} = ReLU(\phi_{l,k})$
        \EndFor
    \EndFor    
    \State \textbf{return} $CV_{M,1}$
\EndProcedure

\Procedure{$Attention$}{$CV_{(1)}, ..., CV_{(X)}$} 
    \For{$i \in 1,..., X$}
        \State $\zeta_i = \frac{\exp({CV_{(i)}})}{\sum_j^{X} \exp(CV_{(j)})}$
    \EndFor    
    \State \textbf{return} $\zeta$
\EndProcedure

\end{algorithmic}
\end{algorithm}

% dataset stats
\begin{table*}[t]
    \centering
    \begin{tabular}{l|c|c|c|c|c|c|c|c|c}
         & \multirow{2}{*}{\#Dialog} & \multirow{2}{*}{\#Utterance} & \multirow{2}{*}{\#Speaker/Dialog} & \multicolumn{2}{c|}{Utterance Len} & \multicolumn{2}{c|}{Vocab} & \multicolumn{2}{c}{Labels}  \\ \cline{5-10}
         & & & & Avg & Max & Hindi & English & Sarcastic & Humorous \\ \hline \hline
         Train & 1100 & 14000 & \multirow{2}{*}{3 (Avg)} & \multirow{2}{*}{20} & \multirow{2}{*}{128} & 27574 & 2462 & 2748 & 5054  \\
         Test & 90 & 1576 & & & & 8664 & 669 & 391 & 740 \\ \hline
    \end{tabular}
    \caption{Statistics of \datasetname\ for code-mixed sarcasm and humor classification. For each utterance, we extract the visual, acoustic, and transcript of the dialog.}
    \label{tab:dataset:stat}
    \vspace{-2mm}
\end{table*}
% dataset stat speaker
\begin{table}[ht!]
    \centering
    \begin{tabular}{l|c|c|c|c|c}
         & \multicolumn{5}{c}{Speakers (Characters)} \\ \cline{2-6}
         & Indravardan & Maya & Saahil & Monisha & Roshesh \\ \hline \hline
         Sarcastic & 1383 & 826 & 692 & 115 & 123 \\
         Humorous & 2391 & 1733 & 733 & 769 & 168 \\ \hline
    \end{tabular}
    \caption{Speaker-wise sarcastic and humorous utterance distribution in \datasetname.}
    \label{tab:dataset:speaker}
    \vspace{-2mm}
\end{table}
\section{Dataset Preparation}
\label{sec:dataset}
Our multi-modal sarcasm and humor classification dataset is based on the video clips of the popular Indian comedy TV show `\textit{Sarabhai vs. Sarabhai}'. The show resolves around the day-to-day life of five family members, namely, Indravardan (aka Indu), Maya, Saahil, Monisha, and Roshesh, with a few infrequent characters. Each scene of the show involves conversation among two or more speakers, and based on the speaker, we split the conversation into utterances. In all, we extract more than 15K utterances from 400 scenes spread across 50 episodes. We refer to the conversation (or sequence of utterances) in each scene as a standalone dialog. For each utterance in the dialog, we assign appropriate sarcasm (\textit{sarcastic} or \textit{non-sarcastic}) and humor (\textit{humorous} or \textit{non-humorous}) labels. 
Thus, the context for any utterance is restricted to the conversation in the current dialog only.
We employed three annotators for assigning sarcasm and humour labels to each utterance. Finally, we aggregate the annotations using majority voting. We also calculate the Cohen Kappa inter-rater agreement score for the annotations. The average score for humor classification is $0.654$, whereas for sarcasm detection it is $0.681$.

\subsection{Data Preprocessing}
The multi-modal information extraction from a comedy video poses two primary challenges: (1) alignment of the multi-modal signals, and (2) laughter removal from the acoustic signal. For the alignment, we mark the boundary of each utterance on the time spectrum for mapping the corresponding speech and visual frames. This was performed by detecting a prolonged silence in the video, and subsequently, discarding the silence portions on the time spectrum. As a consequence, we obtain the boundary for each utterance in the dialog. Subsequently, we extract the speech signals employing the Google Speech API-based automatic speech recognition tool, called Gnani.ai.\footnote{\url{Gnani.ai}}

Like many other comedy shows, our input video also contains audiences' laugh as they react to the scene. It is a popular practice to highlight the comic or humorous situation in the video. Since one of our target tasks is the humor classification, we remove laughter from the audio signal to avoid the model to overfit on the audience laugh. We employ open source Audacity\footnote{\url{https://github.com/audacity/audacity/blob/master/src/effects/NoiseReduction.cpp}} tool for the laughter and background noise removal. Audacity's algorithm works as follows --- It initially identifies  different sound bands corresponding to the laughter frequency range. It then suppresses the audio frequency signals above the threshold of the laughter sample frequency. Then a sampling function is applied to smooth the suppressed audio, resulting in an audio file with reduced laughter frequency bands.

\subsection{Data Statistics}
In Table \ref{tab:dataset:stat}, we list the dataset statistics along with the annotated class label counts. We split the dataset into train and test set with 1,100 and 90 dialogues, respectively. Furthermore, we use $10\%$ of train set as the validation set during experiments. Out of 14,000 utterances in the train set, the number of sarcastic and humorous utterances are 2,748 and 5,054, respectively. Similarly, the test set comprises 391 sarcastic and 740 humorous utterances. Table \ref{tab:dataset:stat} also lists the word distribution for the Hindi-English code-mixed input. \datasetname\ consists of $\sim$36,000 Hindi and  $\sim$3,000 English words.

We also present the speaker-wise sarcastic and humorous statistics in Table \ref{tab:dataset:speaker}. Out of the five speakers, one speaker stands out in both sarcastic and humorous utterances, i.e., Indravardan, followed by Maya, Saahil, and others.  

% results-separate-v2
\begin{table*}[ht!]
    \centering
    \resizebox{\textwidth}{!}{%
    \begin{tabular}{l|l|c|c|c|c||c|c|c|c}
        \multirow{2}{*}{Modality} & \multicolumn{1}{c|}{\multirow{2}{*}{Model}} & \multicolumn{4}{c||}{Sarcasm Detection} & \multicolumn{4}{c}{Humor Classification} \\ \cline{3-10}
         & & Pre & Rec & F1 & Acc & Pre & Rec & F1 & Acc \\ \hline \hline
         \multirow{2}{*}{Acoustic (A)} & LSTM($A$) & 0.419 & 0.146 & 0.216 & 0.738 & 0.475 & 0.139 & 0.215 & 0.523 \\
        &  LSTM(\textit{H-ATN}$^A$) &  0.383 &  0.129 &  0.193 &  0.537 &  0.445 &  0.171 &  0.247 &  0.463 \\
        & LSTM($A$) + \textit{C-ATN}$^D$ & 0.422 & 0.222 & 0.290 & 0.628 & 0.400 & 0.570 & 0.470 & 0.851 \\
        &  LSTM(\textit{H-ATN}$^A$) + \textit{C-ATN}$^D$ &  0.254 &  0.207 &  0.228 &  0.537 &  0.273 &  0.597 &  0.375 &  0.619\\
        \hline
        \multirow{4}{*}{Text (T)} & LSTM($T_{avg}$) & 0.569 & 0.558 & 0.563 & 0.669 & 0.753 & 0.617 & 0.678 & 0.867 \\
        &  LSTM($T_{BERT}$) &  0.646 &  0.524 &  0.579 &  0.774 &  0.707 &  0.667 &  0.687 &  0.717 \\
        & LSTM(\textit{H-ATN}$^U$) & 0.862 & 0.573 & 0.688 & 0.871 & 0.711 & 0.724 & 0.717 & 0.735 \\
        & LSTM(\textit{H-ATN}$^U$) + \textit{C-ATN}$^D$ & 0.833 & 0.601 & 0.698 & 0.871 & 0.760 & 0.830 & 0.793 & 0.797 \\ \hline
        \multirow{2}{*}{Visual (V)}
        & LSTM($V_a$) &  0.113 &  0.084 &  0.096 &  0.178 &  0.161 &  0.105 &  0.127 &  0.226 \\
        &  LSTM($V_g$) &  0.318 &  0.127 &  0.182 &  0.310 &  0.387 &  0.179 &  0.245 &  0.491 \\
        \hline
        \multirow{4}{6em}{Text+Acoustic (T+A)} & LSTM($A$) + LSTM($T_{avg}$) & 0.571 & 0.401 & 0.563 & 0.789 & 0.528 & 0.662 & 0.587 & 0.628 \\
         & LSTM($A$) + LSTM(\textit{H-ATN}$^U$) & 0.801 & 0.586 & 0.674 & 0.865 & \bf 0.809 & 0.801 & 0.805 & 0.818 \\
         & LSTM($A$) + LSTM(\textit{H-ATN}$^U$) + \textit{C-ATN}$^D$ &  \bf 0.865 & 0.555 & 0.676 & 0.868 & 0.755 & 0.832 & 0.797 & 0.807 \\
         & LSTM($A$) + LSTM(\textit{H-ATN}$^U$) + \textit{C-ATN}$^D$ + Filter & 0.811 & \bf 0.636 &  \bf 0.711 & \bf 0.873 & 0.785 & \bf 0.858 & \bf 0.820 & \bf 0.823 \\
         \hline
         \multirow{2}{10em}{{Text+Acoustic+Visual (T+A+V)}}
         & LSTM($V_a$) + LSTM($A$) + LSTM(\textit{H-ATN}$^U$) + \textit{C-ATN}$^D$ + Filter & 0.695 & 0.596 & 0.642 & 0.726 &   0.800 &  0.747 &  0.773 &  0.810\\
        &  LSTM($V_g$) + LSTM($A$) + LSTM(\textit{H-ATN}$^U$) + \textit{C-ATN}$^D$ + Filter &  0.748 &  0.571 &  0.647 &  0.774 &  0.762 &  0.775 &  0.768 &  0.792\\
         \hline
    \end{tabular}%
    }
    \caption{Experimental results for the sarcasm detection and humor classification. All models of each task are separately trained and evaluated. $A \rightarrow$ Acoustic features from Librosa;\textit{ H-ATN}$^U \rightarrow$ Utterance-level hierarchical attention mechanism over textual modality ; \textit{H-ATN}$^A \rightarrow$ Utterance-level hierarchical attention mechanism over acoustic modality; \textit{C-ATN}$^D \rightarrow$ Dialog-level contextual attention mechanism; $T_{avg} \rightarrow$ Textual Utterance embedding computed as an average of the constituents word embeddings; $V_a \rightarrow$ Visual features from Affectiva; $V_g \rightarrow$ Visual features from GoogleNet.}
    \label{tab:results:separate}
    \vspace{-5mm}
\end{table*}

An excerpt from \datasetname\ dataset representing a dialog is presented in Table \ref{tab:error:dialog}. The dialog comprises 10 utterances between two speakers, Maya and Indravardhan. The transcripts are mainly in romanized Hindi with many English words (27 out of total 180 words) in between, such as [\textit{phone}] in utterance $u_1$, [\textit{come}, \textit{on}, \textit{don't} \& \textit{cry}] in $u_4$, [\textit{please}, \textit{you} \& \textit{know}] in $u_5$, [\textit{come}, \textit{on} \& \textit{vegetable}] in $u_6$, and so on. For each utterance, we also report the annotated labels for the sarcasm and humor, i.e., two utterances ($u_4$ and $u_7$) are annotated as sarcastic and four utterances ($u_2, u_3, u_4$ and $u6$) are annotated as humorous.  

\vspace{-3mm}
\subsection{Feature Extraction}
\label{sec:dataset:feature:ex}
We employ pre-trained FastText multilingual word embedding model \cite{conneau2017word,lample2017unsupervised} and Librosa \cite{brian_mcfee-proc-scipy-2015} tool for the textual and acoustic representations, respectively. For each token in the utterance, we extract a 300-dimensional word vector. Following the uncased version, we obtain embedding coverage for more than 90\% of vocabulary words. 
For the acoustic representation, we use Librosa \cite{brian_mcfee-proc-scipy-2015} tool to extract the acoustic features for each frame -- we extract the maximum possible (128) MFCCs (Mel-frequency cepstral coefficients) for every frame. To obtain the utterance-level acoustic representation, we follow the standard acoustic feature extraction technique \cite{abdoli2019end,sharan2019acoustic,khamparia2019sound} by utilizing a time distributed 1D convolution layer on top of MFCCs of all frames.
We do not use visual signals in our models because the quality of visual frames present in our dataset was not good enough, and thus the features extracted from these frames were acting as noise.

\section{Experiments and Analysis} \label{sec:experiment}
\subsection{Experimental Setup} 
We implemented our model in Python-based PyTorch deep learning library. For the evaluation, we compute precision, recall, F1-score, and accuracy for both the tasks. Though we compute and report both accuracy and F1-score for the sake of completeness, our preferred evaluation criteria is F1-score due to the unbalanced label distribution of classification labels (e.g., \textit{sarcastic}/\textit{non-sarcastic}: $391/1185$) in the dataset. We employ forward LSTM \cite{lstm1997} to learn the contextual pattern of the dialog, where each state of the recurrent layer learns a $128$ dimensional hidden vector. We set \textit{dropout}$=40\%$ \cite{dropout}, \textit{batch size}$=32$, and \textit{ReLU} \cite{glorot2011deep:relu} as the activation function for the experiments. At the output, we employ \textit{sigmoid} with \textit{binary cross-entropy} to compute the loss. Subsequently, the computed loss is backpropagated utilizing the \textit{Adam} \cite{adam} optimizer. 

\subsection{Experimental Results}
For the utterance-level localized hierarchical attention mechanism, we experiment with varying attention widths \textit{AtnWidth}$^U$ in the range $[2, 5]$, and obtain \textit{AtnWidth}$^U$ = 3 to be the optimal value. Similarly, for the dialog-level modality-specific attention mechanism, we observe that \textit{AtnWidth}$^D$ = 5 is best suited for the sarcasm and humor classification. We also experimented with visual features. We used the Affectiva API\footnote{\url{https://github.com/cosanlab/affectiva-api-app}} and the GoogleNet Model \cite{szegedy2015going} to obtain the visual expression features. Model \textit{LSTM}($V_a$) and \textit{LSTM}($V_g$) in Table \ref{tab:results:separate} represent the case when only Affectiva and GoogleNet visual features are used for classification respectively. The last row of the table illustrates the case when all  three modalities are used in our model. It can be observed that the results using only visual features are far from satisfactory. This behavior can be attributed to the fact that the video frames present in our dataset have low quality frames.
Table \ref{tab:results:separate} reports the experimental results for both the tasks. It is to be noted that we train and evaluate all the models for both tasks separately.

\subsubsection{Uni-modal evaluation -- Acoustic} 
The first four rows of Table \ref{tab:results:separate} list the results where we classify the utterance employing the acoustic signals only. We obtain accuracies of 73.8\% and 52.3\% using \textit{LSTM}($A$) model for the sarcasm and humor classification, respectively; whereas, F1-scores of 21.6\% and 21.5\% are reported for the two tasks, respectively.
The possible explanation for low F1 score  would be the absence of any semantic entity in the representation -- acoustic feature mainly captures the intensity, excitation mode, pitch, etc. Together with the textual feature, which contains semantic entities (words), the acoustic feature assists the model in leveraging the acoustic variations (e.g., excitement) for sarcasm and humour classification.
Subsequently, we incorporate a dialog-level contextual attention mechanism \textit{C-ATN}$^D$ over the LSTM layer (i.e., \textit{LSTM}($A$) + \textit{C-ATN}$^D$ model) and observe performance improvements of $\sim$8 and $\sim$26 points in F1-scores for the sarcasm detection and humor classification, respectively. The performance difference between the two models for both tasks is primarily due to the reduction in false negatives (and thus improvements in the recall values). Moreover, we credit the improvement to the attention module, which provides crucial assistance to the model in identifying the underrepresented sarcastic and humor classes. In other words, it helps the model to exploit the semantics of the relevant (attended) context in classifying the utterance as sarcastic or humorous.
We also experiment with acoustic feature obtained from utterance-level hierarchical attention module, \textit{H-ATN}$^A$. We observe a performance decrease of $\sim$7 and $\sim$10 in F1 scores for sarcasm and humor classification respectively when we use \textit{H-ATN}$^A$ instead of $A$. Thus, we continue with using $A$ as our acoustic features.
Another important observation suggests that the width of contextual attention (i.e., the number of contextual utterances considered in the attention computation) has an effect on the systems' performance. As we increase the attention width (\textit{AtnWidth}$^D$) beyond five utterances, the performance of the systems begins to degrades. It suggests that the context of sarcasm or humor usually resides in the close proximity of the target utterance - which intuitively follows the real world as both sarcasm and humor lose their effect and relevancy, if delayed for a longer period. On the other hand, smaller \textit{AtnWidth}$^D$ does not offer sufficient context for the model to learn. Hence, for rest of the experiments, we choose \textit{AtnWidth}$^D$=5.

\subsubsection{Uni-modal evaluation -- Textual}
Similar to the acoustic modality, we also perform experiments with only the textual modality. In total, we perform four variants, i.e., \textit{LSTM}($T_{avg}$), \textit{LSTM}(\textit{H-ATN}$^U$), \textit{LSTM}($T_{BERT}$) and \textit{LSTM}(\textit{H-ATN}$^U$) + \textit{C-ATN}$^D$. The first variant is a vanilla LSTM based classification model trained on the textual utterance embeddings - which is computed as the mean of FastText multilingual embeddings of constituent words, and is represented as $T_{avg}$. The second variant,  \textit{LSTM}(\textit{H-ATN}$^U$), is similar to the first except that the textual utterance embeddings is computed utilizing the utterance-level hierarchical attention module. The third variant is also similar to the first two with the difference of type or utterance embedding used. In this variant, we experimented with BERT \cite{devlin2018bert} to get the utterance embeddings. The fourth variant is an extension of the second where we also incorporate the dialog-level contextual attention in classifying the utterances. We evaluate all these variants on both tasks and report the results in the third, fourth, and fifth rows of Table \ref{tab:results:separate}. In the sarcasm detection, we obtain F1-scores of 56.3\%, 68.8\%, and 69.8\% for the three variants, respectively. Similarly, the models yield 67.8\%, 71.7\%, and 79.3\% F1-scores in the humor classification. We can observe that the incorporation of utterance-level hierarchical and dialog-level contextual attention mechanisms have positive effect on the overall performance in both tasks. 

\subsubsection{Bi-modal evaluation -- Textual + Acoustic}
Next, we leverage the availability of both modalities (i.e., text and acoustic) for training \name. We learn two separate LSTMs for each modality, and at each step, we combine the two representations together utilizing the three dialog-level contextual attention modules, i.e., text-specific contextual attention, acoustic-specific contextual attention, and cross-modal contextual attention on both text and acoustic signals. Further, we incorporate a gating mechanism to filter out the noise from the learned representations. Similar to the earlier case, we also experiment with the two variants of the textual representations, i.e., a mean vector $T_{avg}$ and the vector computed by employing hierarchical attention \textit{H-ATN}$^U$. The Text+Acoustic part of Table \ref{tab:results:separate} report the ablation results for the different combinations of individual components - with the last row representing the complete model, as depicted in Figure \ref{fig:arch}. 

The first model under the bi-modal inputs (i.e., \textit{LSTM}($A$)+\textit{LSTM}($T_{avg}$) model) yields 56.3\% and 58.7.0\% F1-scores for the sarcasm and humor classification, respectively. It can be observed that the simple addition of the acoustic information to the textual information does not effect the performance of the system (\textit{LSTM}($A$)+\textit{LSTM}($T_{avg}$)) in a positive way, as compared to the system (\textit{LSTM}($T_{avg}$)) with textual information only (c.f. row three of Table \ref{tab:results:separate}). We observe a performance drop of 9 point in F1-score in humor classification and no changes in case of sarcasm detection. Similarly, we see $\sim$2\% drop in the accuracies values with the simple incorporation of acoustics signal for both tasks. This phenomenon can be attributed to the fact that the two modalities does not complement each other in the feature space and treat each other as the potential noise. We argue that the fusion of two modalities should be performed in an intelligent way such that they complement each other in the model training, and our incorporation of the filtering mechanism in the proposed model, indeed, assists the system to extract the complementary features only.    

The second model, \textit{LSTM}($A$) + \textit{LSTM}(\textit{H-ATN}$^U$), and the third model \textit{LSTM}($A$) +  \textit{LSTM}(\textit{H-ATN}$^U$) + \textit{C-ATN}$^D$) with bi-modal inputs, reflect the incorporation of utterance-level hierarchical and the dialog-level contextual attention modules. However, similar to the previous case, acoustic signal does not have a positive influence on the results for the sarcasm detection.

In the subsequent experiment, we evaluate our complete model on the two tasks, i.e., with the incorporation of filtering mechanism to dictate the complementary feature extraction. The proposed model yields the best F1-scores of 71.1\% and 82.0\% for the sarcasm and humor classification - approximately 4\% and 2\% jump in F1-scores of both sarcasm and humor classification, as compared to the model without filtering mechanism (c.f., second last row of Table \ref{tab:results:separate}). Moreover, it is also evident that the filtering mechanism leverages the acoustic signals in association with the textual information with a $\sim$2\% jump in F1-scores compared to the text-based model. We also observe improvements in accuracy values for the two tasks as well.   

In summary, we observe the following phenomena:
\begin{itemize}[leftmargin=*]
    \item As evident from the obtained results, the textual utterance embedding computed using hierarchical attention mechanism extracts richer features than the mean vector.
    \item The dialog-level contextual attention module learns relevant context to conceive below-the-surface semantic for the target utterance.
    \item The filtering mechanism helps the system to extract relevant information from a modality in the proximity of others.
\end{itemize}

% results-joint-v2
\begin{table*}[ht!]
    \centering
    \resizebox{\textwidth}{!}{%
    \begin{tabular}{l|l|c|c|c|c||c|c|c|c}
        \multirow{2}{*}{Modality} & \multicolumn{1}{c|}{\multirow{2}{*}{Model}} & \multicolumn{4}{c||}{Sarcasm Detection} & \multicolumn{4}{c}{Humor Classification} \\ \cline{3-10}
        & & Pre & Rec & F1 & Acc & Pre & Rec & F1 & Acc  \\ \hline \hline
        \multirow{2}{*}{Acoustic (A)} & LSTM($A$) & 0.457 & 0.107 & 0.174 & 0.747 & 0.543 & 0.291 & 0.379 & 0.552 \\ 
        &  LSTM(\textit{H-ATN}$^A$) &  0.297 &  0.126 &  0.177 &  0.611 &  0.391 &  0.255 &  0.309 &  0.442 \\
        & LSTM($A$) + \textit{C-ATN}$^D$ & 0.520 & 0.263 & 0.350 & 0.757 & 0.565 & 0.381 & 0.455 & 0.572 \\
        &  LSTM(\textit{H-ATN}$^A$) + \textit{C-ATN}$^D$ &  0.432 &  0.211 &  0.255 &  0.635 &  0.498 &  0.356 &  0.415 &  0.557 \\\hline
        \multirow{3}{*}{Text (T)}  & LSTM($T_{avg}$) & 0.803 & 0.488 & 0.607 & 0.843 & 0.749 & 0.818 & 0.782 & 0.786 \\ 
        & LSTM(\textit{H-ATN}$^U$) & \bf 0.843 & 0.506 & 0.633 & 0.854 & 0.754 & 0.832 & 0.791 & 0.794 \\ 
        & LSTM(\textit{H-ATN}$^U$) + \textit{C-ATN}$^D$ & 0.695 & \bf 0.634 & 0.663 & 0.840 & 0.745 & 0.877 & 0.806 & 0.801 \\ \hline
        \multirow{4}{6em}{Text+Acoustic (T+A)}  & LSTM($A$) + LSTM($T_{avg}$) & 0.675 & 0.537 & 0.598 & 0.821 & 0.718 & 0.764 & 0.740 & 0.748 \\ 
        & LSTM($A$) + LSTM(\textit{H-ATN}$^U$) & 0.799 & 0.540 & 0.644 & 0.852 & \bf 0.776 & 0.824 & 0.799 & 0.806 \\ 
        & LSTM($A$) + LSTM(\textit{H-ATN}$^U$) + \textit{C-ATN}$^D$ & 0.800 & 0.552 & 0.654 & 0.855 & 0.766 & 0.851 & 0.807 & 0.808 \\ 
        & LSTM($A$) + LSTM(\textit{H-ATN}$^U$) + \textit{C-ATN}$^D$ + Filter & 0.785 & 0.609 & \bf 0.686 & \bf 0.862 & 0.756 & \bf 0.882 & 0.814 & \bf 0.811 \\
        \hline
    \end{tabular}%
    }
    \caption{Experimental results for the {\em joint-learning} of sarcasm detection and humor classification. \textit{H-ATN}$^U \rightarrow$ Utterance-level hierarchical attention mechanism over textual modality; \textit{C-ATN}$^D \rightarrow$ Dialog-level contextual attention mechanism. $T_{avg}$: Textual utterance embedding computed as an average of the constituents word embeddings.}
    \label{tab:results:joint}
    \vspace{-3.5mm}
\end{table*}

\vspace{-3.5mm}
\subsection{Joint-learning of Sarcasm and Humor}
Since the two tasks are related in the problem space, i.e., both are classification tasks and both have dependencies on the context to extract the hidden semantics, we learn the sarcasm and humor classification tasks in a joint framework. The base architecture (till the filter module) for the joint-learning remains the same as earlier. We only add task-specific layers at the output, i.e., the architecture is extended with two branches corresponding to the two tasks after the filter module. During training, we compute loss at both the branches and propagate them back to the network. 
The results obtained using the joint-learning approach are reported in Table \ref{tab:results:joint}. We repeat the same set of experiments as in the case of separate learning (c.f. Table \ref{tab:results:separate}). 

We can observe that the obtained results follow the same trend as in the case of separate learning. The usage of hierarchical attention, contextual attention, and the filtering modules help the system to obtain the F1-scores of 68.6\% and 81.4\% for the sarcasm detection and humor classification, respectively. However, excluding the filtering module, the system yields inferior F1-scores of 65.4\% and 80.7\%, respectively. Moreover, the incorporation of acoustic information without filtering mechanism also degrades the performance of the system.

In comparison with the separate learning of two tasks, the joint-learning architecture yields lesser performance by 2.5 and 0.6 points in F1-scores; however, it requires lesser (approximately half) parameters to learn, and hence is about 50\% less complex than the two separate models combined.  

\subsection{Comparative Analysis}
We also perform comparative analysis by evaluating the existing systems on \datasetname. In particular, we evaluate \datasetname\ dataset on the following baseline models.
\begin{itemize}[leftmargin=*]
    \item {\bf SVM} \cite{Cortes:1995:SN:218919.218929}: We incorporate an SVM classifier on standalone utterances (without any context) as the baseline system. Depending on the textual representation, we evaluate two variants: a) on the average of the constituent word embeddings ($T_{avg}$), and b) on the embedding computed using the hierarchical attention module \textit{H-ATN}$^U$. For the acoustic signal, we utilize the raw feature representation as mentioned in Section \ref{sec:dataset:feature:ex}.     
    \item {\bf MUStARD} \cite{castro-etal-2019-towards:mustard}: It is an SVM-based system that takes an utterance and its contextual utterances for the classification. For the evaluation, we define previous five utterances as the context and learn the sarcastic and humorous utterance classification. We experiment with the publicly available implementation\footnote{\url{https://github.com/soujanyaporia/MUStARD}} provided by Castro et al. \cite{castro-etal-2019-towards:mustard}.
    \item {\bf Ghosh et al.} \cite{ghosh-veale-2017-magnets}: The underlying architecture of Ghosh et al. \cite{ghosh-veale-2017-magnets} also incorporates the contextual information while classifying an utterance. The authors proposed a deep neural network architecture that models the contextual and target utterances using two separate CNN-BiLSTM layers. Further, the learned representations are combined in DNN for the classification\footnote{Gosh et al. \cite{ghosh-veale-2017-magnets} also employed authors' profile information for the modeling; however, we did not utilize such information during the evaluation.}. Similar to the earlier case, for the evaluation, we define previous five utterances as context. The implementation of the model was adopted from \cite{ghosh-veale-2017-magnets}\footnote{\url{https://github.com/AniSkywalker/SarcasmDetection}}.   
    \item {\bf DialogRNN} \cite{drnn:aaai:erc}: The DialogRNN (DRNN) \cite{drnn:aaai:erc} is one of the recent classification models capable of handling conversational dialog. It was originally proposed for the emotion recognition in conversation (ERC) task; however, it is the closest approach considering our modeling of the two tasks, i.e., classifying each utterance in the conversational dialog. The DRNN architecture encodes speaker-specific utterances independent of other speakers, and subsequently, incorporates each speaker-specific sequence to maintain the dialog sequence. We utilize the implementation\footnote{\url{https://github.com/declare-lab/conv-emotion}} of DRNN \cite{drnn:aaai:erc} for the evaluation.     
\end{itemize}
% results-compare
\begin{table}[ht!]
    \centering
    \resizebox{\columnwidth}{!}{
    \begin{tabular}{l|l|c|c|c|c||c|c|c|c}
        \multirow{2}{*}{} & \multicolumn{1}{c|}{\multirow{2}{*}{Systems}} & \multicolumn{4}{c||}{Sarcasm Detection} & \multicolumn{4}{c}{Humor Classification} \\ \cline{3-10}
         & & Pre & Rec & F1 & Acc & Pre & Rec & F1 & Acc \\ \hline \hline
         \multirow{5}{*}{\rot{Text}} & SVM ($T_{avg}$) & 0.170 & 0.332 & 0.225 & 0.618 & 0.284 & 0.475 & 0.356 & 0.492 \\
          & SVM (\textit{H-ATN}$^U$) & 0.320 & 0.343 & 0.331 & 0.656 & 0.658 & 0.299 & 0.411 & 0.598 \\
         & MUStARD \cite{castro-etal-2019-towards:mustard} & 0.510 & 0.404 & 0.451 & 0.756 & 0.673 & 0.538 & 0.598 & 0.661 \\
         & Ghosh et al. \cite{ghosh-veale-2017-magnets} & 0.595 & 0.432 & 0.500 & 0.786 & 0.648 & 0.518 & 0.576 & 0.644 \\
         & DialogRNN \cite{drnn:aaai:erc} & 0.751 & \bf 0.604 & 0.670 & 0.759 & \bf 0.730 & 0.698 & 0.714 & \bf 0.764 \\
         
         & \textbf{\name} & \bf 0.833 & 0.601 & \bf 0.698 & \bf 0.871 & 0.711 & \bf 0.724 & \bf 0.759 & 0.735 \\ \hline
         
         \multirow{5}{*}{\rot{Text + Audio}} & SVM ($T_{avg}$+\textit{A}) & 0.217 & 0.281 & 0.245 & 0.571 & 0.389 & 0.370 & 0.379 & 0.431 \\
          & SVM (\textit{H-ATN}$^U$+\textit{A}) & 0.274 & 0.384 & 0.320 & 0.595 & 0.429 & 0.397 & 0.413 & 0.469 \\
         & MUStARD \cite{castro-etal-2019-towards:mustard} & 0.520 & 0.458 & 0.487 & 0.761 & 0.692 & 0.546 & 0.610 & 0.673 \\
         & DialogRNN \cite{drnn:aaai:erc} & 0.725 & \bf 0.690 & 0.708 & 0.761 & 0.714 & 0.725 & 0.720 & 0.749 \\
        & \textbf{\name} & \bf 0.853 & 0.636 & \bf 0.711 & \bf 0.873 & \bf 0.785 & \bf 0.858 & \bf 0.820 & \bf 0.823 \\ \hline

    \end{tabular}}
    \caption{Comparative study against existing approaches. \textbf{MUStARD \cite{castro-etal-2019-towards:mustard}:} SVM-based system with pre-defined context; \textbf{Ghosh et al. \cite{ghosh-veale-2017-magnets}:} CNN-BiLSTM with pre-defined context; \textbf{DRNN \cite{drnn:aaai:erc}:} Recurrent model for the classification of each utterance in the conversational dialog.} 
    \label{tab:results:compare}
\end{table}

% error-analysis
\begin{table*}[ht!]
    \centering
    \begin{tabular}{c|l|p{40em}|c|c|c|c}
         \multirow{2}{*}{\#} & \multirow{2}{*}{Speaker} & \multirow{2}{*}{Utterance} & \multicolumn{2}{c|}{Sarcasm} & \multicolumn{2}{c}{Humor} \\ \cline{4-7}
         & & & Actual & Pred & Actual & Pred \\ \hline \hline
        \multirow{2}{*}{$u_1$} & \multirow{2}{*}{ Maya: } & {\em Viren ka \textcolor{blue}{phone} aaya tha, Los Angeles se. mere popat kaka.} \hfill [Eng words: 1] & \multirow{2}{*}{\xmark} & \multirow{2}{*}{\xmark} & \multirow{2}{*}{\xmark} & \multirow{2}{*}{\xmark} \\ \cline{3-3}
         &    &  I got a call from Viren from Los Angeles. My Popat uncle. & &    &    &    \\ \hline
        \multirow{2}{*}{$u_2$} & \multirow{2}{*}{ Indu: } & {\em gaye kya?} \hfill [Eng words: 0] & \multirow{2}{*}{\xmark} & \multirow{2}{*}{\xmark} & \multirow{2}{*}{\checkmark} & \multirow{2}{*}{\checkmark} \\ \cline{3-3}
         &    &  Did he die? &    &    &    &    \\ \hline
        \multirow{2}{*}{$u_3$} & \multirow{2}{*}{ Maya: } & {\em nahin. tayaari mein hain} \hfill [Eng words: 0] & \multirow{2}{*}{\xmark} & \multirow{2}{*}{\xmark} & \multirow{2}{*}{\textcolor{red}{\checkmark}} & \multirow{2}{*}{\textcolor{red}{\xmark}} \\ \cline{3-3}
         &    &  No, preparing for it. &    &    &    &    \\ \hline
        \multirow{2}{*}{$u_4$} & \multirow{2}{*}{ Indu: } & {\em \textcolor{blue}{come on come on}, maya. \textcolor{blue}{don’t cry}. tum janti ho tum roti ho aur bhi acchi nahin lagti} \textcolor{white}{.........}\hfill [Eng words: 6] & \multirow{2}{*}{\checkmark} & \multirow{2}{*}{\checkmark} & \multirow{2}{*}{\checkmark} & \multirow{2}{*}{\checkmark} \\ \cline{3-3}
         &    & Come on Come on, Maya. Don't cry. You know when you cry, you look even worse. &    &    &    &    \\ \hline
        \multirow{2}{*}{$u_5$} & \multirow{2}{*}{ Maya: } & {\em Indravardan! \textcolor{blue}{Please! you know}, viren keh raha tha ki unhone bilkul bistar pakad liya hai. chal phir bhi nahi sakte bechare} \hfill [Eng words: 3] & \multirow{2}{*}{\xmark} & \multirow{2}{*}{\xmark} & \multirow{2}{*}{\textcolor{red}{\xmark}} & \multirow{2}{*}{ \textcolor{red}{\checkmark} } \\ \cline{3-3}
         &    & Indravardan! Please! you know, Viren was saying he is completely bed-ridden, the miserable man can't even walk. &    &    &    &    \\ \hline
        \multirow{2}{*}{$u_6$} & \multirow{2}{*}{ Indu: } & {\em \textcolor{blue}{come on} maya vo navve saal ke hain. is umr mein koi bhi insaan \textcolor{blue}{vegetable} jaisa ho jaata hai} \hfill [Eng words: 3] & \multirow{2}{*}{\textcolor{red}{\xmark}} & \multirow{2}{*}{ \textcolor{red}{\checkmark} } & \multirow{2}{*}{\checkmark} & \multirow{2}{*}{\checkmark} \\ \cline{3-3}
         &    & Come on Maya. He is 90 years old. At his age, every one becomes miserable (seems like a vegetable).  &    &    &    &    \\ \hline
        \multirow{2}{*}{$u_7$} & \multirow{2}{*}{ Maya: } & {\em \textcolor{blue}{I know, I know, darling. I mean}, apni monisha ko hi dekh lo, itni choti umr mein bilkul \textcolor{blue}{vegetable} jaisi ho gayi hai. din bhar \textcolor{blue}{sofa} par padi rehti hai. dopahar ki \textcolor{blue}{t.v. serial} dekhti rehti hai. sabzi bhi wahin lete lete kaatti hai. chai piti hai to \textcolor{blue}{glass} uthakar phir \textcolor{blue}{kitchen} nahin le rakhti. tel laga sir \textcolor{blue}{sofa} ke \textcolor{blue}{cushions} mein rakh deti hai} \hfill [Eng words: 15] & \multirow{2}{*}{\checkmark} & \multirow{2}{*}{\checkmark} & \multirow{2}{*}{\xmark} & \multirow{2}{*}{\xmark} \\ \cline{3-3}
         &    &  I know, I know, darling. I mean, look at our Monisha, she looks so miserable at this young age. She spend her whole day on the sofa watching daytime T.V. serial. She chops vegetable while reclining there. She does not put the cup back in the kitchen after having a cup of tea. She puts her oily hair on sofa's cushion. &    &    &    &    \\ \hline
        \multirow{2}{*}{$u_8$} & \multirow{2}{*}{ Indu: } & {\em popat kaka. maya, hum popat kaka ki baat kar rahe the, na} \hfill[Eng words: 0] & \multirow{2}{*}{\xmark} & \multirow{2}{*}{\xmark} & \multirow{2}{*}{\xmark} & \multirow{2}{*}{\xmark} \\ \cline{3-3}
         &    &  Popat uncle! Maya, were'nt we talking about Popat uncle?  &    &    &    &    \\ \hline
        \multirow{2}{*}{$u_9$} & \multirow{2}{*}{ Maya: } & {\em haan, viren keh raha tha ki din bhar mujhe yaad karte rehte hain. maya, maya, maya ko bulao} \hfill [Eng words: 0] & \multirow{2}{*}{\xmark} & \multirow{2}{*}{\xmark} & \multirow{2}{*}{\textcolor{red}{\xmark}} & \multirow{2}{*}{ \textcolor{red}{\checkmark} } \\ \cline{3-3}
         &    &  Yes, Viren was mentioning that he remembers me the whole day. Maya, Maya, ask Maya to come. &    &    &    &    \\ \hline
        \multirow{2}{*}{$u_{10}$} & \multirow{2}{*}{ Indu: } & {\em accha to kab jana ho raha hai los angeles tumhara?} \hfill [Eng words: 0] & \multirow{2}{*}{\xmark} & \multirow{2}{*}{\xmark} & \multirow{2}{*}{\xmark} & \multirow{2}{*}{\xmark} \\ \cline{3-3}
         &    &  Great, when are you leaving for Los Angeles?  &    &    &    &    \\ \hline
    \end{tabular}
    \caption{Actual and predicted labels for sarcasm detection and humor classification for a dialog having 10 utterances ($u_1, ..., u_{10}$) in \datasetname\ dataset. Blue-colored texts represent English words, while black-colored texts are either romanized Hindi or named entities. For sarcasm detection, \name\ yields 66\% precision and 100\% recall. Similarly, we obtain 60\% precision and 75\% recall for the humor classification.}
    \label{tab:error:dialog}
    \vspace{-5mm}
\end{table*}

In Table \ref{tab:results:compare}, we report the results of above comparative systems. For each comparative system, we evaluate on both uni-modal\footnote{We do not report uni-modal \textit{acoustic} results due to extremely poor performance.} \textit{textual} and bi-modal \textit{textual+acoustic} information. In text modality, SVM on $T_{avg}$ reports mediocre F1-scores of 22.5\% and 35.6\% for the sarcasm and humor classification, respectively. In contrast, the same SVM classifier improves the performance of two tasks (11\% and 6\%, respectively) by utilizing the embeddings of hierarchical attention module. In comparison, the contextual models (MUStARD \cite{castro-etal-2019-towards:mustard} and Ghosh et al. \cite{ghosh-veale-2017-magnets}) yield decent F1-scores of 45.1\% and 50.0\% in sarcasm detection. Similarly, the two comparative systems obtains 59.8\% and 57.6\% F1-scores for the humor classification. Finally, we evaluate DialogRNN \cite{drnn:aaai:erc} for both sarcasm and humor classification, and obtains the best comparative F1-scores of 67.0\% and 71.4\%, respectively. In comparison, for the same input (i.e., textual modality), our proposed system reports $\sim$3\% and $\sim$4.5\% improvement over the best comparative system.  

We observe similar trends with the bi-modal \textit{textual+acoustic} inputs for both the tasks under consideration. The SVM-based system records the least F1-scores of 24.5\% and 37.9\%, while DialogRNN \cite{drnn:aaai:erc} reports the best performance among the comparative systems with 70.8\% and 72.0\% F1-scores for the sarcasm and humor classification tasks, respectively. Comparison shows the superiority of the proposed system over the comparative system with $>$1 and 10 points improvement in the F1-scores.    
     
\subsection{Error Analysis}
Though \name\ performs better than the existing systems, it did misclassify some utterances as well. In this section, we report our quantitative and qualitative analysis on the errors. At first, we analyze the system's performance in terms of confusion matrix, as depicted in Table \ref{tab:confusion}.

\begin{table}[t]
    \centering
    \begin{tabular}{l|c|c|}
         \multicolumn{1}{c}{} & \multicolumn{1}{c}{Sar} & \multicolumn{1}{c}{Non-Sar} \\ \cline{2-3}
         Sar & 249 & 142 \\ \cline{2-3}
         Non-Sar & 58 & 1127 \\ \cline{2-3}
    \end{tabular}\hspace{2em}
    \begin{tabular}{l|c|c|}
         \multicolumn{1}{c}{} & \multicolumn{1}{c}{Hum} & \multicolumn{1}{c}{Non-Hum} \\ \cline{2-3}
         Hum & 635 & 105 \\ \cline{2-3}
         Non-Hum & 174 & 662 \\ \cline{2-3}
    \end{tabular}
    \caption{Confusion matrix for \name.}
    \label{tab:confusion}
\end{table}

% heatmaps
\begin{figure*}
    \centering
    \subfloat[Textual attention. \label{fig:heatmap:text:humour}]{
    \includegraphics[height=7em]{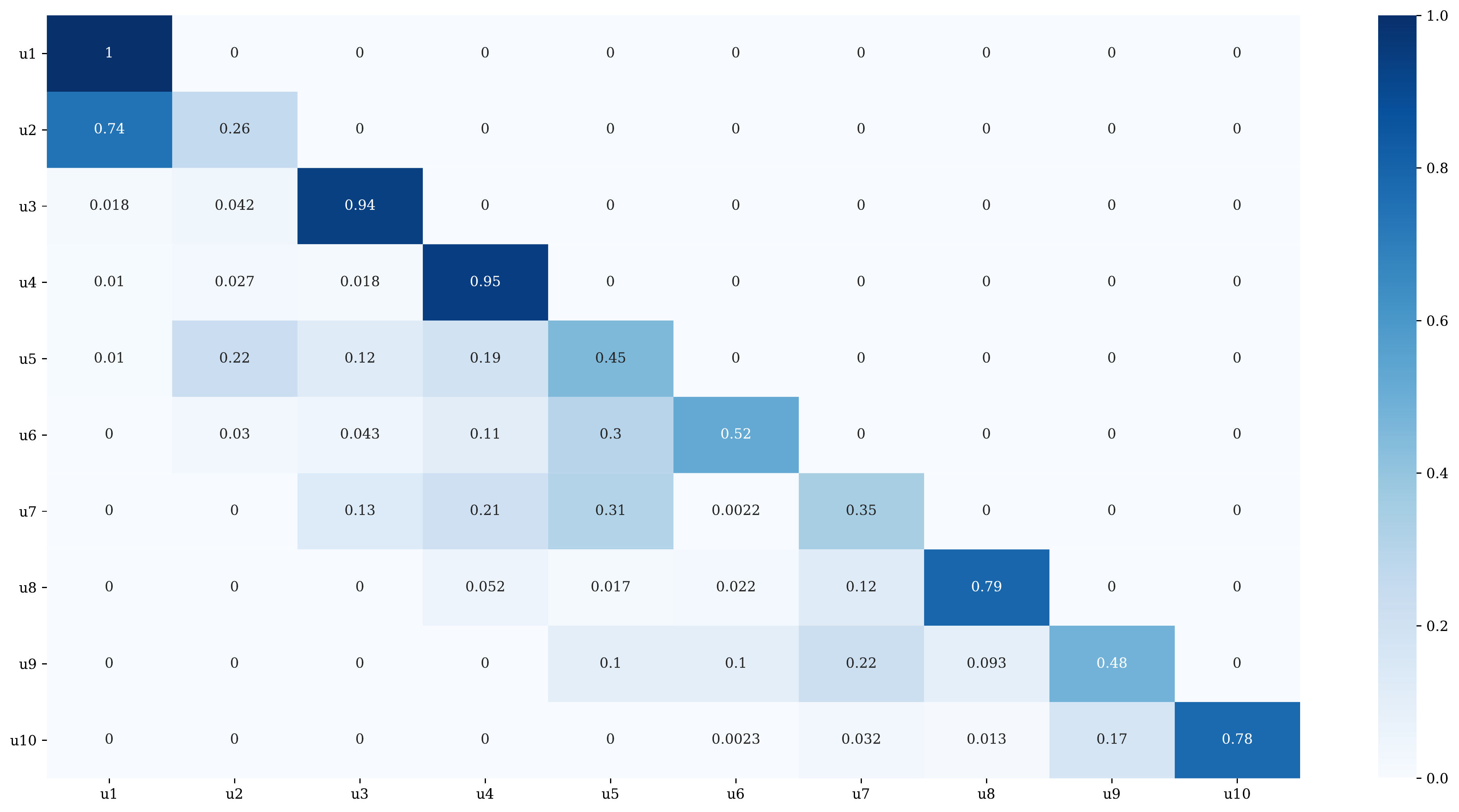}}
    \hspace{1em}
    \subfloat[Acoustic attention.\label{fig:heatmap:audio:humour}]{
    \includegraphics[height=7em]{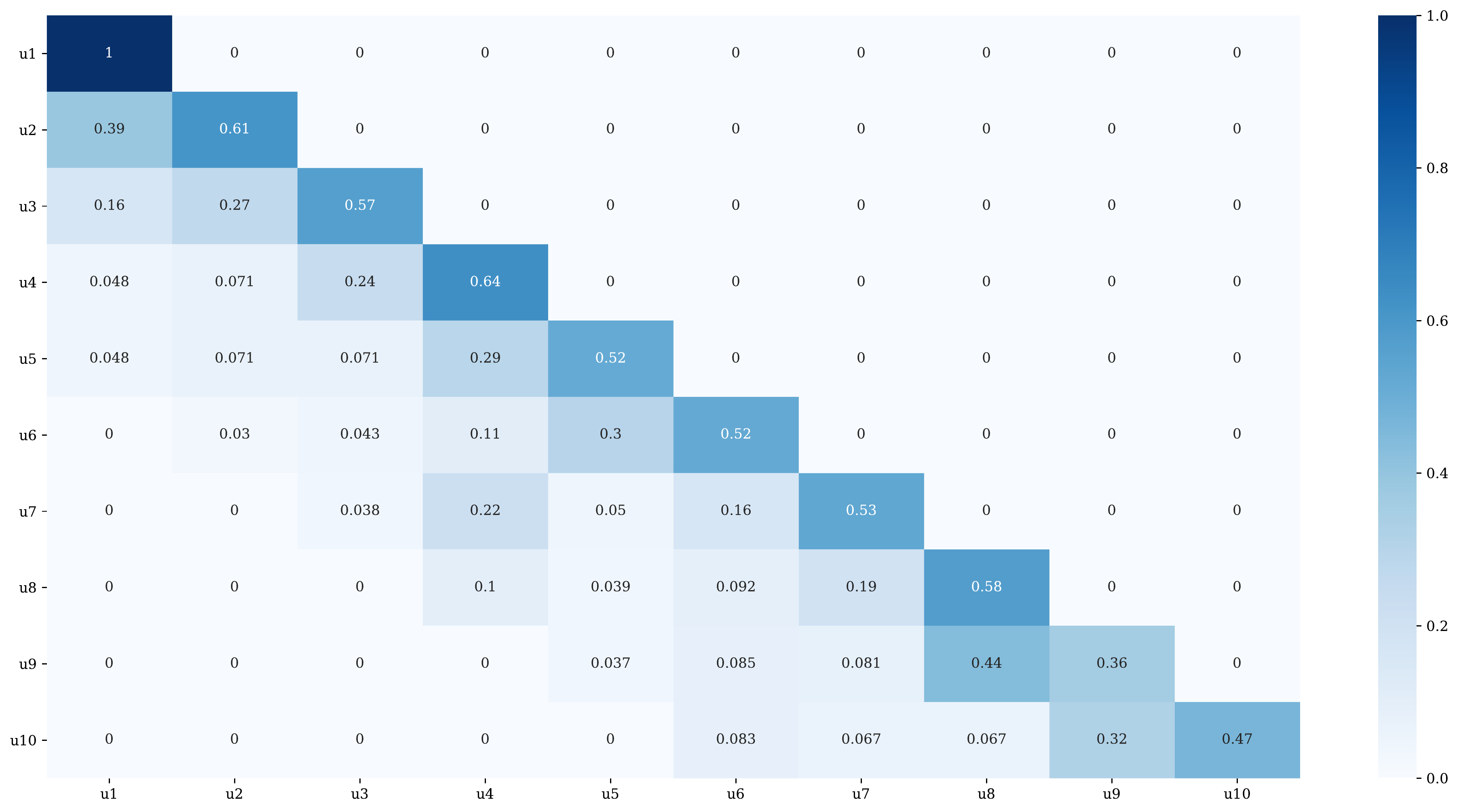}}
    \hspace{1em}
    \subfloat[Textual and acoustic cross-modal attention.\label{fig:heatmap:text:audio:humour}]{
    \includegraphics[width=0.44\textwidth, height=7em]{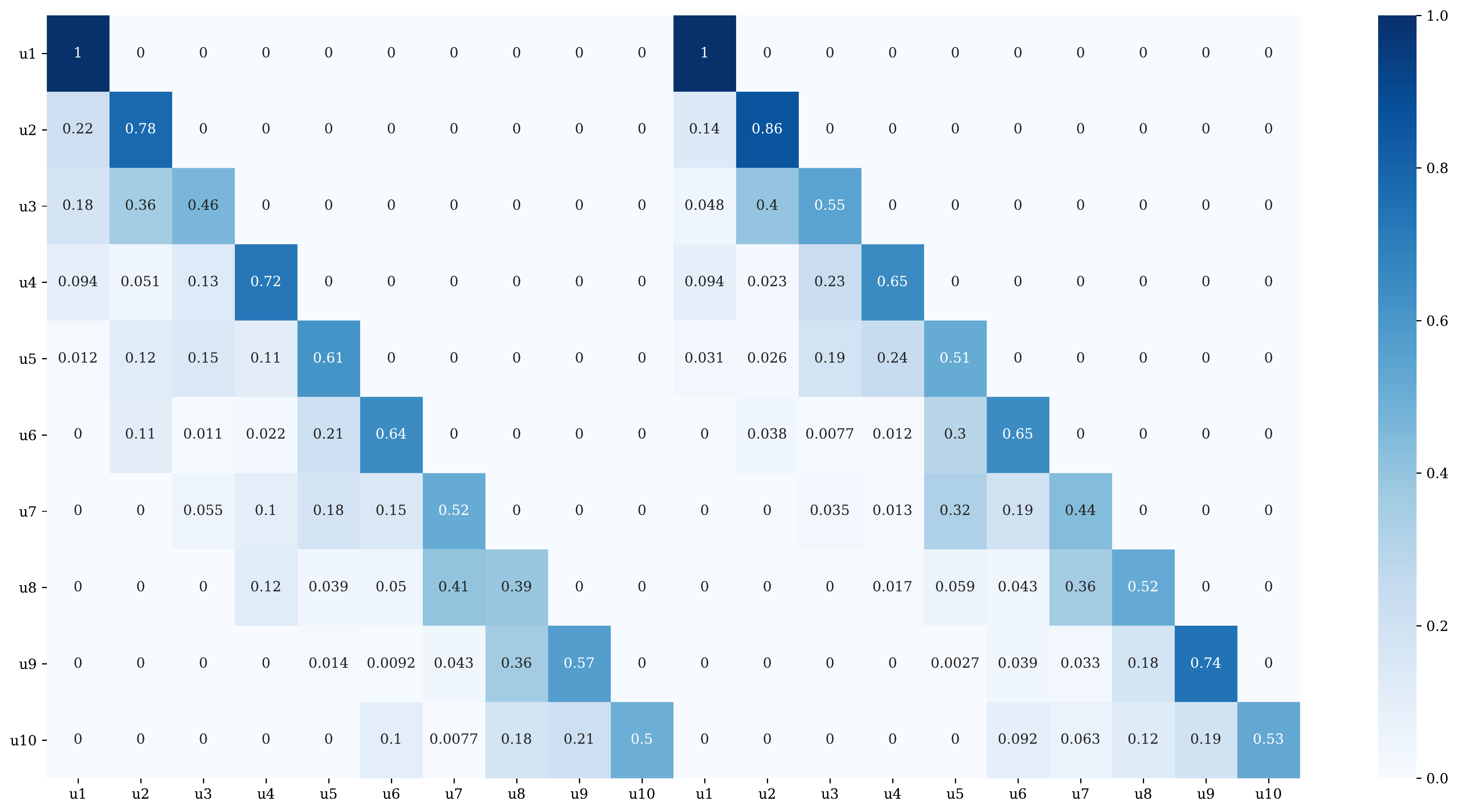}} 
    
    \caption{Humor Classification: Heatmap analysis of the dialog-level contextual attention module for the dialog presented in Table \ref{tab:error:dialog}. For each utterance $u_i$ on the y-axis, we compute attention weights for the 5 utterances, i.e., the current and previous four utterances (\textit{AtnWidth}$^D$ = 5). The cell values $(i,i-4), (i,i-3), (i,i-2), (i,i-1),$ and $(i,i)$ represents the attention weights of utterances $u_{i-4}, u_{i-3}, u_{i-2}, u_{i-1},$ and $u_{i}$, respectively. The colormap signifies the amount attention weight for the respective utterances. The darker the shade, higher the attention weight assigned by \name.}
    \label{fig:heatmap:humor}
\end{figure*}

\begin{figure*}
    \centering
    \subfloat[Textual attention.\label{fig:heatmap:text:sarcasm}]{
    \includegraphics[ height=7em]{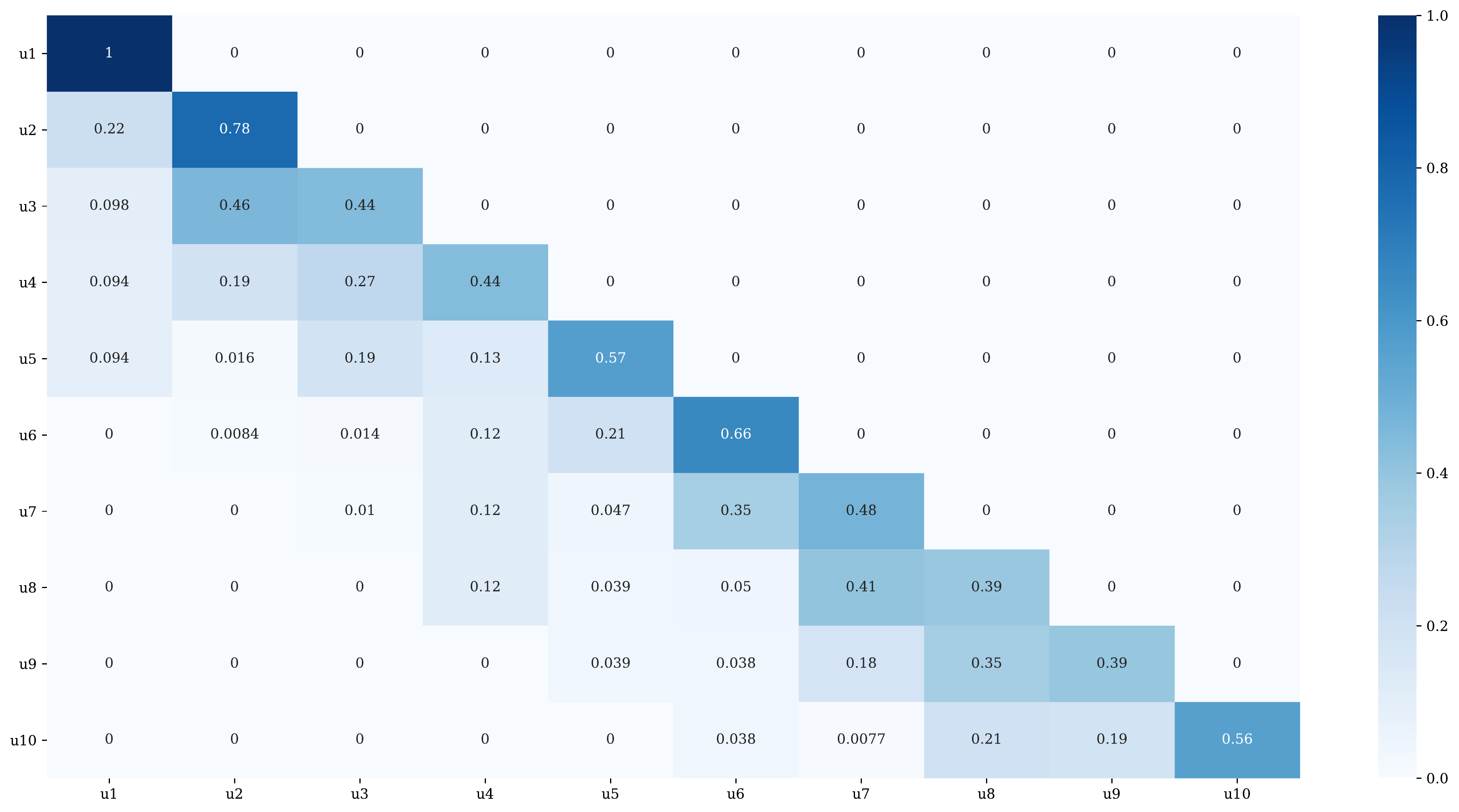}}
    \hspace{1em}
    \subfloat[Acoustic attention.\label{fig:heatmap:audio:sarcasm}]{
    \includegraphics[ height=7em]{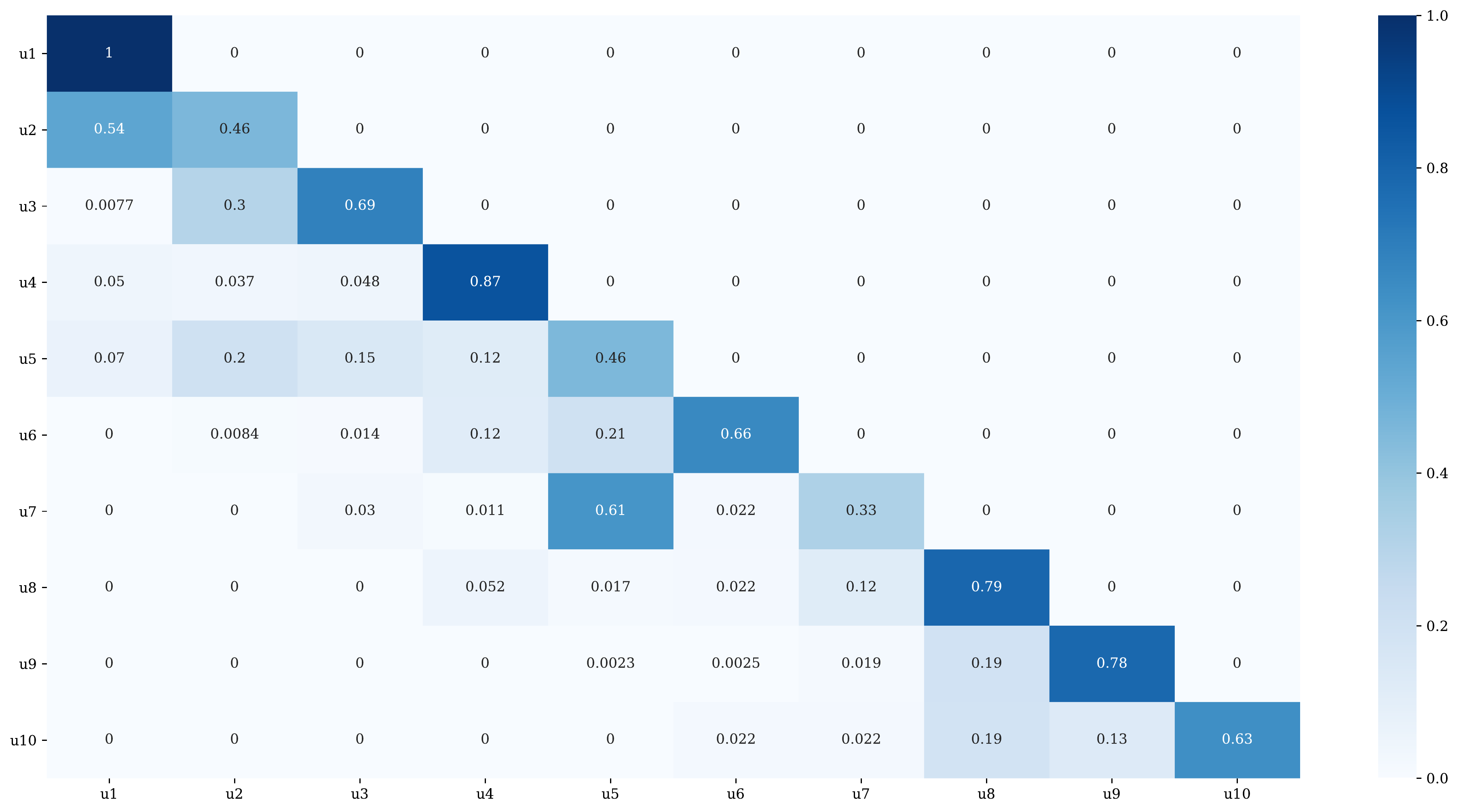}}
    \hspace{1em}
    \subfloat[Textual and acoustic cross-modal attention.\label{fig:heatmap:text:audio:sarcasm}]{
    \includegraphics[width=0.44\textwidth, height=7em]{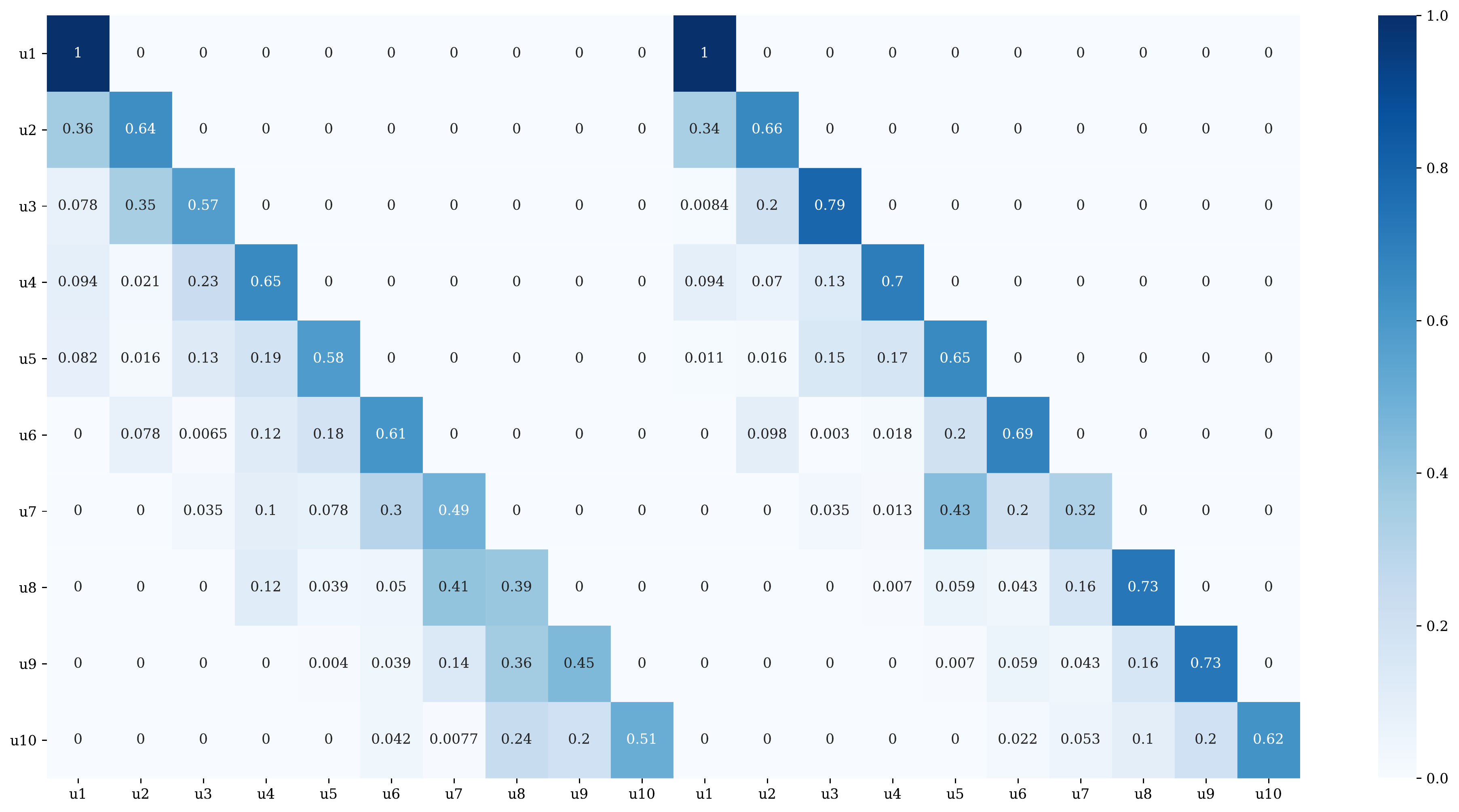}} 

    \caption{Sarcasm Detection: Heatmap   of the dialog-level contextual attention module for the dialog presented in Table \ref{tab:error:dialog}.}
    \label{fig:heatmap:sarcasm}
    
\end{figure*}

Next, we choose a dialog (consisting of 10 utterances) from the test set and present system's prediction in Table \ref{tab:error:dialog}. It reports code-mixed utterances (with English translation), its speakers, and its actual and predicted labels for both sarcasm and humor classification tasks. Across 10 utterances in the dialog, two of them are labeled as sarcastic in the gold set, while, the count of humorous utterances is 4. 
The dialog in Table \ref{tab:error:dialog} also exhibits the contextual and/or multi-modal dependencies for an utterance to be labeled as sarcastic/humorous. For example, humorous utterances $u_2$ and $u_3$ do not convey any explicit textual semantics on their own; instead, they rely on the contextual utterances, i.e., $u_1$ for $u_2$ and $u_1 \& u_2$ for $u_3$. Moreover, utterance $u_2$ also depends on the multi-modal information, i.e., the excited voice of the speaker (Indu) along with the context signals the presence of humor in $u_2$. We also highlight the English words (bold text) along with its count for each utterance in the dialog which constitute approximately 15\% of the complete text. Out of these English words, some word plays crucial role in the identification of sarcasm/humor in the utterance. For example, the metaphorically used English word `\textit{vegetable}' in utterance $u_6$ is the prime clue for the utterance to be identified as humorous.

On evaluation, our system predicts three utterances as sarcastic (i.e., $u_4, u_7,$ and $u_7$) and 5 utterances as humorous (i.e., $u_2, u_4, u_5, u_6,$ and $u_9$). In both cases, it makes some correct predictions as well as some incorrect predictions. For the sarcasm detection, our system yields \textit{precision} of 66\% (i.e., two out of three predictions are correct) and \textit{recall} of 100\% (i.e., both sarcastic utterances are correctly predicted). On the other hand, we obtain \textit{precision} and \textit{recall} of 60\% and 75\%, respectively, for the humor classification, i.e., we observe two \textit{false-positives} and one \textit{false-negatives} along with three \textit{true-positives}.

We also analyse the effect of the level of code-mixing for both the tasks. In the given example, there is on an average one English word in every six words. For sarcasm classification, our model only returns one false positive. It is for the case where the utterance contains three English words out of the total 19 words, i.e., one English word for every 6.3 words. Whereas it gives correct predictions for utterances 4 and 7 which contains one English word for every three words and one English word for every 4.3 words, respectively. For the humor classification task, our model misclassifies for utterances u3, u5 and u9,  having 0, 3 and 0 English words respectively. The ratio for English words to total words then turns out to be 0:4, 1:3 and 0:18. Whereas it predicts humor correctly for utterances u2, u4 and u6, having 0, 6 and 3 English words respectively. The ratio of English words to total being 0:2, 1:3 and 1:6.3 respectively. Looking at these results, we hypothesize that our model performs better for the utterances containing comparatively more English words.
To validate this hypothesis, we sample two sets from our test set. Both the set has 100 utterances. One of the set contains the utterances having the most number of English words ($\sim$18 English words per utterance) and another contains utterances containing the least number of English words ($\sim$1 English word per utterance).
We evaluate our final model on these two sets and report the result in Table \ref{tab:code-mix-ablation}. It can be easily seen from the table that the our model performs better when there are more English words in the utterance thus validating our hypothesis.
% code-mix-ablation
\begin{table}[t]
\centering
\resizebox{\columnwidth}{!}{%
\begin{tabular}{l|l|l|l|l|l|l|l|l|l}
\multicolumn{1}{c|}{\multirow{2}{*}{Model}} & \multicolumn{1}{c|}{\multirow{2}{*}{Data}} & \multicolumn{4}{c|}{Sarcasm Detection} & \multicolumn{4}{c}{Humor Classification} \\ \cline{3-10} 
\multicolumn{1}{c|}{} & \multicolumn{1}{c|}{} & \multicolumn{1}{c|}{Prec} & \multicolumn{1}{c|}{Rec} & \multicolumn{1}{c|}{F1} & \multicolumn{1}{c|}{Acc} & \multicolumn{1}{c|}{Prec} & \multicolumn{1}{c|}{Rec} & \multicolumn{1}{c|}{F1} & \multicolumn{1}{c}{Acc} \\ \hline \hline
\multirow{2}{*}{\textbf{\name}} & \textit{Set1} & \bf 0.67 & \bf 0.79 & \bf 0.72 & \bf 0.83 & \bf 0.92 & 0.91 & \bf 0.91 & \bf 0.87 \\ \cline{2-10} 
 & \textit{Set2} & 0.62 & 0.62 & 0.62 & 0.82 & 0.51 & \bf 0.96 & 0.67 & 0.76 \\ \hline
\end{tabular}%
}
\caption{Experimental results on sampled data from test set to analyze the effect of the extent of code-mixing in our model. \textit{Set1} contains 100 utterances containing $\sim$18 English words (on avg) per sentence, whereas \textit{Set2} contains 100 utterances with $\sim$1 English word per sentence.}
\label{tab:code-mix-ablation}
\end{table}

We also perform heatmap analysis of the attention weights as computed by the system. For the analysis, we take the same dialog as presented in Table \ref{tab:error:dialog} and depict the heatmaps of  dialog-level contextual attention \textit{C-ATN}$^D$ in Figures \ref{fig:heatmap:humor} and \ref{fig:heatmap:sarcasm} for the humor classification and sarcasm detection, respectively. For each case, we show three separate heatmaps of the attention matrices corresponding to the textual, acoustics, and cross-modal attention modules. Each row $i$ represents an utterance for which we compute attention weights for the five (current and four previous) utterances (i.e., \textit{AtnWidth}$^D$=5) and the color shade signifies the amount of attention the model assigns to the corresponding utterances - darker shades represent higher weight, while lighter shades signify lower weights. Rest of the entries have attention weight zero, as they don't participate in the computation.

From the heatmaps, we can observe that the attention modules assign different wights to the contextual utterances depending upon their importance. For example, the system assigns higher weight on the previous textual content (c.f. Figure \ref{fig:heatmap:text:humour} $\beta_1=0.74 \& \beta_2=0.26$) and the current acoustic context (c.f. Figure \ref{fig:heatmap:audio:humour} $\alpha_1=0.39 \& \alpha_2=0.61$) for utterance $u_2$ in the humor classification. The distribution of attention weights can be justified by manual observation as well. The textual content of $u_2$ (i.e., \textit{gaye kya?}|Did he die?) does not offer significant information about being humorous and one has to consider the context for the semantic. On the other hand, the audio signal reveals the excitement and tone in the voice of the speaker, and thus validates the higher attention weights by the system. Similarly, we observe many scenarios in other dialogs as well where the attention weights have high correlation with the contextual and multi-modal semantics of the utterances.

\section{Conclusion and Future Work}
\label{sec:con}
In this paper, we presented our research on the Hindi-English code-mixed conversational dialog. We developed \datasetname, a qualitative multi-modal dataset for the sarcasm detection and humor classification in code-mixed conversations. The utterances in our dataset is adopted from a popular Indian television comedy show. We collected, cleaned, and annotated more than 15000 utterances across 1190 dialogs. To evaluate \datasetname, we proposed \name, an attention-based multi-modal classification model for the utterance classification. To learn an enriched textual representation for the utterance, we proposed and implemented an utterance-level hierarchical attention module. Evaluation results showed the incorporation of enriched textual representation has a positive impact on the performance. Further, we benchmark our \datasetname\ dataset on \name\ by performing comparative analysis against exiting classification approaches.      

In the light of recent advancements in representational learning, in future, we would like to explore the multi-lingual embeddings for the efficient code-mixed representations. 

\section*{Acknowledgment}
The work was partially supported by the Ramanujan Fellowship (SERB) and the Infosys Centre for AI, IIITD.

{\small
\bibliography{affect-ref}

\begin{thebibliography}{10}
\providecommand{\url}[1]{#1}
\csname url@rmstyle\endcsname
\providecommand{\newblock}{\relax}
\providecommand{\bibinfo}[2]{#2}
\providecommand\BIBentrySTDinterwordspacing{\spaceskip=0pt\relax}
\providecommand\BIBentryALTinterwordstretchfactor{4}
\providecommand\BIBentryALTinterwordspacing{\spaceskip=\fontdimen2\font plus
\BIBentryALTinterwordstretchfactor\fontdimen3\font minus
  \fontdimen4\font\relax}
\providecommand\BIBforeignlanguage[2]{{%
\expandafter\ifx\csname l@#1\endcsname\relax
\typeout{** WARNING: IEEEtran.bst: No hyphenation pattern has been}%
\typeout{** loaded for the language `#1'. Using the pattern for}%
\typeout{** the default language instead.}%
\else
\language=\csname l@#1\endcsname
\fi
#2}}

\bibitem{Turney:2002}
P.~D. Turney, ``Thumbs up or thumbs down?: Semantic orientation applied to
  unsupervised classification of reviews,'' in \emph{ACL}, 2002, pp. 417--424.

\bibitem{Panget.al2005}
B.~Pang, , and L.~Lee, ``{Seeing Stars: Exploiting Class Relationships for
  Sentiment Categorization with respect to Rating Scales},'' in \emph{ACL},
  2005, pp. 115--124.

\bibitem{taffc:2016:sentiment}
C.~{Clavel} and Z.~{Callejas}, ``Sentiment analysis: From opinion mining to
  human-agent interaction,'' \emph{IEEE Transactions on Affective Computing},
  vol.~7, no.~1, pp. 74--93, 2016.

\bibitem{senti:lstm:cambria:2018}
Y.~Ma, H.~Peng, T.~Khan, E.~Cambria, and A.~Hussain, ``{Sentic LSTM: a Hybrid
  Network for Targeted Aspect-Based Sentiment Analysis},'' \emph{Cognitive
  Computation}, vol.~10, pp. 639--650, 2018.

\bibitem{ekman1999}
P.~Ekman, \emph{Basic Emotions}.\hskip 1em plus 0.5em minus 0.4em\relax The
  handbook of cognition and emotion., 1999.

\bibitem{taffc:2019:multitask:acoustic:emotion}
B.~{Zhang}, E.~M. {Provost}, and G.~{Essl}, ``Cross-corpus acoustic emotion
  recognition with multi-task learning: Seeking common ground while preserving
  differences,'' \emph{IEEE Transactions on Affective Computing}, vol.~10,
  no.~1, pp. 85--99, 2019.

\bibitem{akhtar:all:in:one:affective}
M.~S. Akhtar, D.~Ghosal, A.~Ekbal, P.~Bhattacharyya, and S.~Kurohashi,
  ``{All-in-One: Emotion, Sentiment and Intensity Prediction using a Multi-task
  Ensemble Framework},'' \emph{IEEE Transactions on Affective Computing}, pp.
  1--1, 2019.

\bibitem{akhtar:cim:emotion:2020}
M.~S. Akhtar, A.~Ekbal, and E.~Cambria, ``{How Intense Are You? Predicting
  Intensities of Emotions and Sentiments using Stacked Ensemble},'' \emph{IEEE
  CIM}, vol.~15, pp. 64--75, 2020.

\bibitem{kumar2021discovering}
S.~Kumar, A.~Shrimal, M.~S. Akhtar, and T.~Chakraborty, ``Discovering emotion
  and reasoning its flip in multi-party conversations using masked memory
  network and transformer,'' \emph{arXiv preprint arXiv:2103.12360}, 2021.

\bibitem{cai-etal-2019-multi}
Y.~Cai, H.~Cai, and X.~Wan, ``Multi-modal sarcasm detection in twitter with
  hierarchical fusion model,'' in \emph{ACL}, 2019, pp. 2506--2515.

\bibitem{castro-etal-2019-towards:mustard}
S.~Castro, D.~Hazarika, V.~P{\'e}rez-Rosas, R.~Zimmermann, R.~Mihalcea, and
  S.~Poria, ``Towards multimodal sarcasm detection (an {\_}{O}bviously{\_}
  perfect paper),'' in \emph{ACL}, 2019, pp. 4619--4629.

\bibitem{suyash:multimodal:sarcasm:ijcnn:2020}
S.~Sangwan, M.~S. Akhtar, P.~Behera, and A.~Ekbal, ``{I didn't mean what I
  wrote! Exploring Multimodality for Sarcasm Detection},'' in \emph{IJCNN},
  2020.

\bibitem{hasan-etal-2019-ur}
M.~K. Hasan, W.~Rahman, A.~Bagher~Zadeh, J.~Zhong, M.~I. Tanveer, L.-P.
  Morency, and M.~E. Hoque, ``{UR}-{FUNNY}: A multimodal language dataset for
  understanding humor,'' in \emph{EMNLP-IJCNLP}, 2019, pp. 2046--2056.

\bibitem{humor:text:audion:lrec:2016}
D.~Bertero and P.~Fung, ``Deep learning of audio and language features for
  humor prediction,'' in \emph{LREC}, 2016, pp. 496--501.

\bibitem{multimodal1}
S.~Poria, E.~Cambria, D.~Hazarika, N.~Majumder, A.~Zadeh, and L.-P. Morency,
  ``Context-dependent sentiment analysis in user-generated videos,'' in
  \emph{ACL}, vol.~1, 2017, pp. 873--883.

\bibitem{ghosal-EtAl:2018:EMNLP}
D.~Ghosal, M.~S. Akhtar, D.~Chauhan, S.~Poria, A.~Ekbal, and P.~Bhattacharyya,
  ``{Contextual Inter-modal Attention for Multi-modal Sentiment Analysis},'' in
  \emph{EMNLP}, 2018, pp. 3454--3466.

\bibitem{zadeh2018acl}
A.~Zadeh, P.~P. Liang, S.~Poria, E.~Cambria, and L.-P. Morency, ``Multimodal
  language analysis in the wild: Cmu-mosei dataset and interpretable dynamic
  fusion graph,'' in \emph{ACL}, 2018, pp. 2236--2246.

\bibitem{akhtar:naacl:2019:multitask:multimodal}
M.~S. Akhtar, D.~Chauhan, D.~Ghosal, S.~Poria, A.~Ekbal, and P.~Bhattacharyya,
  ``{Multi-task Learning for Multi-modal Emotion Recognition and Sentiment
  Analysis},'' in \emph{NAACL-HLT}, 2019, pp. 370--379.

\bibitem{akhtar:multimodal:tkdd:2020}
M.~S. Akhtar, D.~S. Chauhan, and A.~Ekbal, ``{A Deep Multi-Task Contextual
  Attention Framework for Multi-modal Affect Analysis},'' \emph{TKDD}, vol.~14,
  pp. 32:1--32:27, 2020.

\bibitem{taffc:2019:audio:visual:emotion}
F.~{Noroozi}, M.~{Marjanovic}, A.~{Njegus}, S.~{Escalera}, and
  G.~{Anbarjafari}, ``Audio-visual emotion recognition in video clips,''
  \emph{IEEE Transactions on Affective Computing}, vol.~10, no.~1, pp. 60--75,
  2019.

\bibitem{conneau2017word}
A.~Conneau, G.~Lample, M.~Ranzato, L.~Denoyer, and H.~J{\'e}gou, ``Word
  translation without parallel data,'' \emph{arXiv preprint arXiv:1710.04087},
  2017.

\bibitem{lample2017unsupervised}
G.~Lample, A.~Conneau, L.~Denoyer, and M.~Ranzato, ``Unsupervised machine
  translation using monolingual corpora only,'' \emph{arXiv preprint
  arXiv:1711.00043}, 2017.

\bibitem{lstm1997}
S.~Hochreiter and J.~Schmidhuber, ``{Long short-term memory},'' \emph{Neural
  computation}, vol.~9, pp. 1735--1780, 1997.

\bibitem{kreuz-caucci:2007:sarcasm:lexical}
R.~Kreuz and G.~Caucci, ``Lexical influences on the perception of sarcasm,'' in
  \emph{Proceedings of the Workshop on Computational Approaches to Figurative
  Language}, 2007, pp. 1--4.

\bibitem{davidov:sarcasm:semisupervised:2010}
D.~Davidov, O.~Tsur, and A.~Rappoport, ``Semi-supervised recognition of
  sarcastic sentences in twitter and amazon,'' in \emph{CoNLL}, 2010, p.
  107–116.

\bibitem{tsur:sarcasm:2010}
O.~Tsur, D.~Davidov, and A.~Rappoport, ``{ICWSM - A Great Catchy Name:
  Semi-Supervised Recognition of Sarcastic Sentences in Online Product
  Reviews},'' in \emph{ICWSM}, W.~W. Cohen and S.~Gosling, Eds., 2010.

\bibitem{joshi:sarcsam:incongruity:2015}
A.~Joshi, V.~Sharma, and P.~Bhattacharyya, ``Harnessing context incongruity for
  sarcasm detection,'' in \emph{ACL}, 2015, pp. 757--762.

\bibitem{sarcasm:twitter:pattern:ieee:access:2016}
M.~{Bouazizi} and T.~{Otsuki Ohtsuki}, ``A pattern-based approach for sarcasm
  detection on twitter,'' \emph{IEEE Access}, vol.~4, pp. 5477--5488, 2016.

\bibitem{joshi:sarcasm:survey:2017}
A.~Joshi, P.~Bhattacharyya, and M.~J. Carman, ``Automatic sarcasm detection: A
  survey,'' \emph{ACM Comput. Surv.}, vol.~50, 2017.

\bibitem{sarcasm:soft:attention:ieee:access:2019}
L.~H. {Son}, A.~{Kumar}, S.~R. {Sangwan}, A.~{Arora}, A.~{Nayyar}, and
  M.~{Abdel-Basset}, ``Sarcasm detection using soft attention-based
  bidirectional long short-term memory model with convolution network,''
  \emph{IEEE Access}, vol.~7, pp. 23\,319--23\,328, 2019.

\bibitem{joshi:sarcasm:history:2015}
A.~Khattri, A.~Joshi, P.~Bhattacharyya, and M.~Carman, ``Your sentiment
  precedes you: Using an author{'}s historical tweets to predict sarcasm,'' in
  \emph{Proceedings of the 6th Workshop on Computational Approaches to
  Subjectivity, Sentiment and Social Media Analysis}, 2015, pp. 25--30.

\bibitem{ghosh-veale-2017-magnets}
A.~Ghosh and T.~Veale, ``Magnets for sarcasm: Making sarcasm detection timely,
  contextual and very personal,'' in \emph{EMNLP}, 2017, pp. 482--491.

\bibitem{hazarika-etal-2018-cascade}
D.~Hazarika, S.~Poria, S.~Gorantla, E.~Cambria, R.~Zimmermann, and R.~Mihalcea,
  ``{CASCADE}: Contextual sarcasm detection in online discussion forums,'' in
  \emph{COLING}, 2018, pp. 1837--1848.

\bibitem{ghosh-etal-2017-role}
D.~Ghosh, A.~Richard~Fabbri, and S.~Muresan, ``The role of conversation context
  for sarcasm detection in online interactions,'' in \emph{Proceedings of the
  18th Annual {SIG}dial Meeting on Discourse and Dialogue}, 2017, pp. 186--196.

\bibitem{DBLP:journals/corr/abs-1805-11869}
S.~Swami, A.~Khandelwal, V.~Singh, S.~S. Akhtar, and M.~Shrivastava, ``A corpus
  of english-hindi code-mixed tweets for sarcasm detection,'' \emph{CoRR}, vol.
  abs/1805.11869, 2018.

\bibitem{sarcasm:hindi:2017}
S.~K. Bharti, K.~Sathya~Babu, and S.~K. Jena, ``Harnessing online news for
  sarcasm detection in hindi tweets,'' in \emph{Pattern Recognition and Machine
  Intelligence}, 2017, pp. 679--686.

\bibitem{mihalcea-strapparava-2005-making}
R.~Mihalcea and C.~Strapparava, ``Making computers laugh: Investigations in
  automatic humor recognition,'' in \emph{EMNLP}, 2005, pp. 531--538.

\bibitem{semeval:2017:humor:potash-etal}
P.~Potash, A.~Romanov, and A.~Rumshisky, ``{S}em{E}val-2017 task 6:
  {\#}{H}ashtag{W}ars: Learning a sense of humor,'' in \emph{SemEval}, 2017,
  pp. 49--57.

\bibitem{Yang2019}
Z.~Yang, B.~Hu, and J.~Hirschberg, ``{Predicting Humor by Learning from
  Time-Aligned Comments},'' in \emph{Interspeech}, 2019, pp. 496--500.

\bibitem{annamoradnejad2021colbert}
I.~Annamoradnejad, ``Colbert: Using bert sentence embedding for humor
  detection,'' 2021.

\bibitem{Humor:Prosody}
A.~Purandare and D.~Litman, ``Humor: Prosody analysis and automatic recognition
  for f*r*i*e*n*d*s*.'' 01 2006, pp. 208--215.

\bibitem{bertero2016predicting}
D.~Bertero and P.~Fung, ``Predicting humor response in dialogues from tv
  sitcoms,'' in \emph{ICASSP}, 2016, pp. 5780--5784.

\bibitem{vaswani:transformer:attention:all:you:need:2017}
A.~Vaswani, N.~Shazeer, N.~Parmar, J.~Uszkoreit, L.~Jones, A.~N. Gomez,
  L.~Kaiser, and I.~Polosukhin, ``{Attention is All you Need},'' in
  \emph{NIPS}, 2017, pp. 5998--6008.

\bibitem{mfn:zadeh:aaa1:2017}
A.~Zadeh, P.~P. Liang, N.~Mazumder, S.~Poria, E.~Cambria, and L.~Morency,
  ``Memory fusion network for multi-view sequential learning,'' in \emph{AAAI},
  2018, pp. 5634--5641.

\bibitem{khandelwal-etal-2018-humor}
A.~Khandelwal, S.~Swami, S.~S. Akhtar, and M.~Shrivastava, ``Humor detection in
  {E}nglish-{H}indi code-mixed social media content : Corpus and baseline
  system,'' in \emph{LREC}, 2018.

\bibitem{sane-etal-2019-deep}
S.~R. Sane, S.~Tripathi, K.~R. Sane, and R.~Mamidi, ``Deep learning techniques
  for humor detection in {H}indi-{E}nglish code-mixed tweets,'' in
  \emph{Proceedings of the Tenth Workshop on Computational Approaches to
  Subjectivity, Sentiment and Social Media Analysis}, 2019, pp. 57--61.

\bibitem{brian_mcfee-proc-scipy-2015}
{B}rian {M}c{F}ee, {C}olin {R}affel, {D}awen {L}iang, {D}aniel
  {P}.{W}.~{E}llis, {M}att {M}c{V}icar, {E}ric {B}attenberg, and {O}riol
  {N}ieto, ``librosa: {A}udio and {M}usic {S}ignal {A}nalysis in {P}ython,'' in
  \emph{{P}roceedings of the 14th {P}ython in {S}cience {C}onference}, 2015,
  pp. 18 -- 24.

\bibitem{he2016deep}
K.~He, X.~Zhang, S.~Ren, and J.~Sun, ``{Deep Residual Learning for Image
  Recognition},'' in \emph{CVPR}, 2016, pp. 770--778.

\bibitem{arevalo:2017:gmu:gated}
J.~Arevalo, T.~Solorio, M.~Montes-y G{\'o}mez, and F.~A. Gonz{\'a}lez, ``Gated
  multimodal units for information fusion,'' \emph{arXiv preprint
  arXiv:1702.01992}, 2017.

\bibitem{abdoli2019end}
S.~Abdoli, P.~Cardinal, and A.~L. Koerich, ``End-to-end environmental sound
  classification using a 1d convolutional neural network,'' \emph{Expert
  Systems with Applications}, vol. 136, pp. 252--263, 2019.

\bibitem{sharan2019acoustic}
R.~V. Sharan and T.~J. Moir, ``Acoustic event recognition using cochleagram
  image and convolutional neural networks,'' \emph{Applied Acoustics}, vol.
  148, pp. 62--66, 2019.

\bibitem{khamparia2019sound}
A.~Khamparia, D.~Gupta, N.~G. Nguyen, A.~Khanna, B.~Pandey, and P.~Tiwari,
  ``Sound classification using convolutional neural network and tensor deep
  stacking network,'' \emph{IEEE Access}, vol.~7, pp. 7717--7727, 2019.

\bibitem{dropout}
N.~Srivastava, G.~Hinton, A.~Krizhevsky, I.~Sutskever, and R.~Salakhutdinov,
  ``{Dropout: A Simple Way to Prevent Neural Networks from Overfitting},''
  \emph{JMLR}, vol.~15, pp. 1929--1958, 2014.

\bibitem{glorot2011deep:relu}
X.~Glorot, A.~Bordes, and Y.~Bengio, ``{Deep Sparse Rectifier Neural
  Networks},'' in \emph{AISTATS}, 2011, p. 315–323.

\bibitem{adam}
D.~P. Kingma and J.~Ba, ``{Adam: A Method for Stochastic Optimization},''
  \emph{CoRR}, vol. abs/1412.6980, 2014.

\bibitem{szegedy2015going}
C.~Szegedy, W.~Liu, Y.~Jia, P.~Sermanet, S.~Reed, D.~Anguelov, D.~Erhan,
  V.~Vanhoucke, and A.~Rabinovich, ``Going deeper with convolutions,'' in
  \emph{CVPR}, 2015, pp. 1--9.

\bibitem{devlin2018bert}
J.~Devlin, M.-W. Chang, K.~Lee, and K.~Toutanova, ``Bert: Pre-training of deep
  bidirectional transformers for language understanding,'' \emph{arXiv preprint
  arXiv:1810.04805}, 2018.

\bibitem{Cortes:1995:SN:218919.218929}
C.~Cortes and V.~Vapnik, ``Support-vector networks,'' \emph{Mach. Learn.},
  vol.~20, pp. 273--297, 1995.

\bibitem{drnn:aaai:erc}
N.~Majumder, S.~Poria, D.~Hazarika, R.~Mihalcea, A.~F. Gelbukh, and E.~Cambria,
  ``Dialoguernn: An attentive {RNN} for emotion detection in conversations,''
  in \emph{AAAI}, 2019, pp. 6818--6825.

\end{thebibliography}
\bibliographystyle{IEEEtran}}

% biography
\begin{wrapfigure}{l}{2cm}
\centering
\includegraphics[width=2cm,height=2cm]{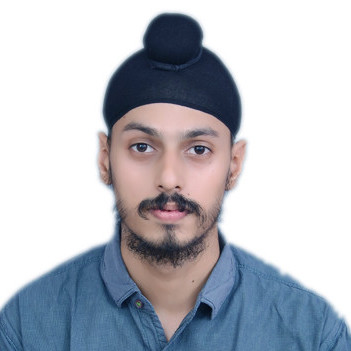}
\end{wrapfigure} 

\noindent \textbf{Manjot Bedi} Manjot Bedi is a Research Associate at IIIT-Delhi. Prior to this, he worked as a Software Developer at Wipro Technologies. His primary research interests include Natural Language Processing, Computer Vision and Machine Learning.\\

\begin{wrapfigure}{l}{2cm}
\centering
\includegraphics[width=2cm,height=2cm]{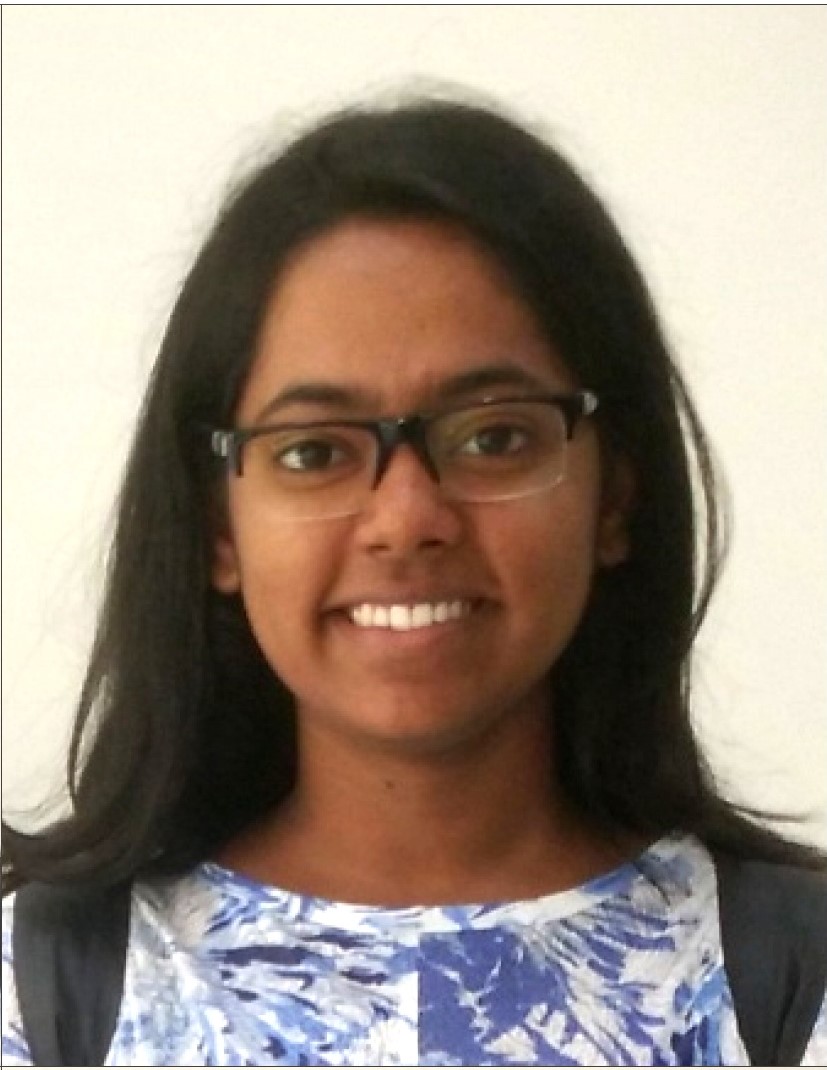}
\end{wrapfigure} 

\noindent \textbf{Shivani Kumar} is a PhD scholar at IIIT-Delhi, India. She holds a Junior Research Fellowship and works in the domain of Natural Language Processing, primarily in the area of conversational analysis. \\

\begin{wrapfigure}{l}{2cm}
\centering
\includegraphics[width=2cm,height=2cm]{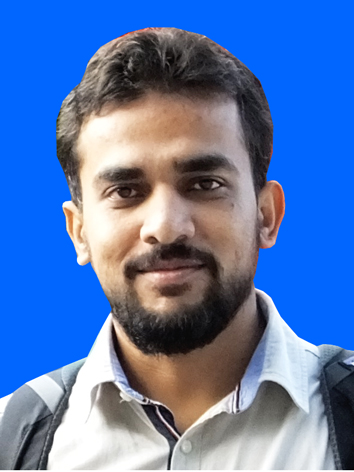}
\end{wrapfigure} 

\noindent \textbf{Md Shad Akhtar} is currently an Assistant Professor at IIIT Delhi. His main area of research is NLP with a focus on the affective analysis. He completed his PhD form IIT Patna.\\

\begin{wrapfigure}{l}{2cm}
\centering
\includegraphics[width=2cm,height=2cm]{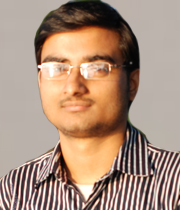}
\end{wrapfigure} 

\noindent \textbf{Tanmoy Chakraborty} is an Assistant Professor and a Ramanujan Fellow at IIIT Delhi, India. Prior to this, he was a postdoctoral researcher at  University of Maryland, College Park, USA. He completed his PhD as a Google India PhD scholar from IIT Kharagpur, India. His primary research interests include Social Computing and NLP.
\end{document}